\documentclass[journal]{IEEEtran}
\usepackage{amsmath,amsfonts}
\usepackage{array}
\usepackage{textcomp}
\usepackage{stfloats}
\usepackage{url}
\usepackage{verbatim}
\usepackage{graphicx}
\usepackage{cite}
\usepackage{multirow}

\usepackage{algpseudocode}
\usepackage{algorithm} 

\usepackage[dvipsnames]{xcolor}

\usepackage[nolist]{acronym}
\usepackage{glossaries}

\usepackage{caption} %
\usepackage{subcaption} %

\usepackage[binary-units]{siunitx}

\usepackage[export]{adjustbox}

\usepackage{amsfonts}
\usepackage{amssymb}
\usepackage{bbm}

\usepackage{wrapfig}
\usepackage{trfsigns}
\usepackage[pdf]{pstricks}
\usepackage{svg}

\usepackage{lipsum}  

\makeglossaries

\newglossaryentry{set_lines_gt}{
  name        = {\ensuremath{G}},
  description = {Menge der Ground Truth Linienzüge},
  sort        = {g}
}

\newglossaryentry{line_gt}{
  name        = {\ensuremath{g}},
  description = {Ground Truth Linienzug},
  sort        = {g}
}

\newglossaryentry{set_points_gt}{
  name        = {\ensuremath{G_{\gls{line_gt}}}},
  description = {Menge der GT Punkte im Linienzug \gls{line_gt}},
  sort        = {g}
}

\newglossaryentry{point_gt}{
  name        = {\ensuremath{p_{\gls{line_gt}}}},
  description = {Punkt im Ground Truth Linienzug \gls{line_gt}},
  sort        = {p}
}

\newglossaryentry{segment_gt}{
  name        = {\ensuremath{s_{\gls{line_gt}}}},
  description = {Segment im Ground Truth Linienzug \gls{line_gt}},
  sort        = {s}
}

\newglossaryentry{class_gt}{
  name        = {\ensuremath{\mathbf{k}_{\gls{line_gt}}}},
  description = {Klasse des Ground Truth Linienzugs \gls{line_gt} als One-Hot-Encoding},
  sort        = {k}
}

\newglossaryentry{set_classes_gt}{
  name        = {\ensuremath{K}},
  description = {Menge der möglichen Klassen},
  sort        = {k}
}

\newglossaryentry{gt_set_segments_grid}{ 
  name        = {\ensuremath{\hat{L}}},
  description = {Menge der GT-Liniensegmente im Zellgitter \gls{set_cells}},
  sort        = {l}
}

\newglossaryentry{gt_segment_grid}{
  name        = {\ensuremath{\mathbf{\hat{\gls{segment_grid}}}}},
  description = {GT-Liniensegment im Zellgitter \gls{set_cells}},
  sort        = {l}
}
                   
\newglossaryentry{gt_set_segments_cell}{
  name        = {\ensuremath{\hat{\gls{set_preds}}}},
  description = {Menge der GT-Liniensegmente je Zelle},
  sort        = {p}
}

\newglossaryentry{set_segments_grid}{ 
  name        = {\ensuremath{L}},
  description = {Menge der Liniensegmente im Zellgitter \gls{set_cells}},
  sort        = {l}
}

\newglossaryentry{preds_matched}{ 
  name        = {\ensuremath{\gls{set_segments_grid}_{\theta}}},
  description = {Menge der Liniensegmente im Zellgitter \gls{set_cells} mit Assoziation},
  sort        = {lt}
}

\newglossaryentry{segment_grid}{
  name        = {\ensuremath{\mathbf{\ell}}},
  description = {geschätztes Liniensegment im Zellgitter \gls{set_cells}},
  sort        = {l}
}

\newglossaryentry{set_cells}{
  name        = {\ensuremath{Z}},
  description = {Menge der Zellen},
  sort        = {z}
}

\newglossaryentry{cell}{
  name        = {\ensuremath{z}},
  description = {Zelle},
  sort        = {z}
}

\newglossaryentry{row}{
  name        = {\ensuremath{r}},
  description = {Zeile im Zellgitter},
  sort        = {r}
}
\newglossaryentry{col}{
  name        = {\ensuremath{o}},
  description = {Spalte im Zellgitter},
  sort        = {o}
}

\newglossaryentry{grid_height}{
  name        = {\ensuremath{h_{\gls{set_cells}}}},
  description = {Höhe des Zellgitters},
  sort        = {h}
}

\newglossaryentry{grid_width}{
  name        = {\ensuremath{w_{\gls{set_cells}}}},
  description = {Breite des Zellgitters},
  sort        = {w}
}

\newglossaryentry{mean_num_lines_img}{
  name        = {\ensuremath{|\overline{\gls{set_segments_grid}}|}},
  description = {Durchschnittliche Anzahl der Liniensegmente je Bild},
  sort        = {l}
}

\newglossaryentry{geom_cart}{
  name        = {\ensuremath{se}},
  description = {Kartesische Koordinaten},
  sort        = {s}
}

\newglossaryentry{pred_start}{
  name        = {\ensuremath{\mathbf{b}}},
  description = {Startpunkt in Zellkoordinaten},
  sort        = {b}
}

\newglossaryentry{pred_end}{
  name        = {\ensuremath{\mathbf{e}}},
  description = {Endpunkt in Zellkoordinaten},
  sort        = {e}
}

\newglossaryentry{geom_md}{
  name        = {\ensuremath{md}},
  description = {Mittelpunkt-Richtung-Koordinaten},
  sort        = {m}
}

\newglossaryentry{pred_midp}{
  name        = {\ensuremath{\mathbf{m}}},
  description = {Mittelpunkt in Zellkoordinaten},
  sort        = {m}
}

\newglossaryentry{pred_dir}{
  name        = {\ensuremath{\mathbf{d}}},
  description = {Differenz zwischen Start-/Endpunkt in Zellkoordinaten},
  sort        = {d}
}

\newglossaryentry{geom_1d}{
  name        = {\ensuremath{1D}},
  description = {1D-Randkoordinaten},
  sort        = {1}
}
\newglossaryentry{pred_1d_start}{
  name        = {\ensuremath{b_{\gls{geom_1d}}}},
  description = {Startpunkt in Randkoordinaten},
  sort        = {b}
}

\newglossaryentry{pred_1d_end}{
  name        = {\ensuremath{e_{\gls{geom_1d}}}},
  description = {Enpunkt in Randkoordinaten},
  sort        = {e}
}

\newglossaryentry{geom_eu}{
  name        = {\ensuremath{eu}},
  description = {Euler-Koordinaten},
  sort        = {e}
}

\newglossaryentry{pred_eu_start}{
  name        = {\ensuremath{\alpha}},
  description = {Start-Winkel in Euler-Koordinaten},
  sort        = {a}
}

\newglossaryentry{pred_eu_end}{
  name        = {\ensuremath{\beta}},
  description = {End-Winkel in Euler-Koordinaten},
  sort        = {b}
}

\newglossaryentry{set_preds}{
  name        = {\ensuremath{P}},
  description = {Menge der Prädiktoren je Zelle},
  sort        = {p}
}
    
\newglossaryentry{pred}{
  name        = {\ensuremath{p}},
  description = {Prädiktor},
  sort        = {p}
}

\newglossaryentry{num_pred}{
  name        = {\ensuremath{|P|}},
  description = {Anzahl an Prädiktoren},
  sort        = {p}
}
\newglossaryentry{num_vars}{
  name        = {\ensuremath{|V|}},
  description = {Anzahl der Trainingsvariablen e.g. $\gls{num_pred}\cdot(4+|\gls{set_classes_gt}+1)|$},
  sort        = {v}
}

\newglossaryentry{geom}{
name        = {\ensuremath{g}},
description = {Geometrieindikator},
sort        = {g}
}

\newglossaryentry{class}{
  name        = {\ensuremath{k}},
  description = {Klassenindikator},
  sort        = {k}
}

\newglossaryentry{pred_geom}{
    name        = {\ensuremath{\mathbf{\gls{geom}}}},
    description = {Geometrie},
    sort        = {g}
}

\newglossaryentry{pred_class}{
  name        = {\ensuremath{\mathbf{\gls{class}}}},
  description = {Klasse als Wahrscheinlichkeitsverteilung},
  sort        = {k}
}

\newglossaryentry{pred_conf}{
  name        = {\ensuremath{c}},
  description = {Konfidenz in YOLinO},
  sort        = {c}
}

\newglossaryentry{pred_conf_gnn}{
  name        = {\ensuremath{\tilde{\gls{pred_conf}}_i}},
  description = {Konfidenz eines Knoten im GNN},
  sort        = {c}
}

\newglossaryentry{pred_conf_gnn_edge}{
  name        = {\ensuremath{\tilde{\gls{pred_conf}}_{ij}}},
  description = {Konfidenz einer Kante im GNN},
  sort        = {c}
}

\newglossaryentry{conf_threshold}{
  name        = {\ensuremath{\tau}},
  description = {Konfidenz-Schwellwert},
  sort        = {t}
}

\newglossaryentry{img_height}{
  name        = {\ensuremath{h}},
  description = {Höhe des Eingangsbildes},
  sort        = {h}
}

\newglossaryentry{img_width}{
  name        = {\ensuremath{w}},
  description = {Breite des Eingangsbildes},
  sort        = {w}
}

\newglossaryentry{loc_loss}{ 
  name = {\ensuremath{\mathcal{L}_{\gls{geom}}}},
  description={Lokalisierungskosten},
  sort        = {lg}
}

\newglossaryentry{conf_loss}{
  name        = {\ensuremath{\mathcal{L}_{\gls{pred_conf}}}},
  description = {Konfidenzkosten},
  sort        = {lc}
}

\newglossaryentry{class_loss}{
  name        = {\ensuremath{\mathcal{L}_{\gls{class}}}},
  description = {Klassifikationskosten},
  sort        = {lk}
}

\newglossaryentry{weight_class}{
  name        = {\ensuremath{\omega_{\gls{class}}}},
  description = {Gewicht der Klassifikationskosten},
  sort        = {wk}  
}

\newglossaryentry{weight_conf}{
  name        = {\ensuremath{\omega_{\gls{pred_conf}}}},
  description = {Gewicht der Konfidenzkosten},
  sort        = {wc}  
}

\newglossaryentry{weight_conf0}{
  name        = {\ensuremath{\omega_{0}}},
  description = {Gewicht der Konfidenzkosten für nicht-verantwortliche Prädiktoren},
  sort        = {wc0}  
}

\newglossaryentry{weight_conf1}{
  name        = {\ensuremath{\omega_{1}}},
  description = {Gewicht der Konfidenzkosten für verantwortliche Prädiktoren},
  sort        = {wc1}  
}

\newglossaryentry{weight_geom}{
  name        = {\ensuremath{\omega_{\gls{geom}}}},
  description = {Gewicht der Geometriekosten},
  sort        = {wg}  
}

\newglossaryentry{graph}{
  name        = {\ensuremath{\mathcal{G}}},
  description = {Graph},
  sort        = {g}
}

\newglossaryentry{edges}{
  name        = {\ensuremath{\mathcal{E}}},
  description = {Menge der Kanten im Graph},
  sort        = {e}
}

\newglossaryentry{edge}{
  name        = {\ensuremath{e}},
  description = {Kante im Graph},
  sort        = {e}
}

\newglossaryentry{nodes}{
  name        = {\ensuremath{\mathcal{V}}},
  description = {Menge der Knoten im Graph},
  sort        = {v}
}

\newglossaryentry{node}{
  name        = {\ensuremath{v}},
  description = {Ein Knoten im Graph},
  sort        = {v}
}

\newglossaryentry{distance}{
    name        = {\ensuremath{\delta}},
    description = {Distanzfunktion},
    sort        = {d}
  }

\newglossaryentry{dbscan_eps}{
    name        = {\ensuremath{\epsilon}},
    description = {Distanzschwellwert in DBSCAN},
    sort        = {e}
  }                 
\newglossaryentry{dbscan_min_pts}{
    name        = {\ensuremath{p_{min}}},
    description = {Minimale Anzahl der Punkte in DBSCAN-Nachbarschaft},
    sort        = {p}
  }                 

\newglossaryentry{nms_mp}{
  name        = {\ensuremath{\mathbf{m}_{uv}}},
  description = {Mittelpunkt in uv-Koordinaten},
  sort        = {m}
}               

\newglossaryentry{nms_dir}{
  name        = {\ensuremath{\mathbf{d}_{uv}}},
  description = {Richtung in uv-Koordinaten},
  sort        = {d}
}                  

\newglossaryentry{nms_len}{
  name        = {\ensuremath{l_{uv}}},
  description = {Länge in uv-Koordinaten},
  sort        = {l}
}                                                                 
\definecolor{my-green}{RGB}{0,100,60}
\definecolor{my-yellow}{RGB}{250,187,8}
\definecolor{my-red}{RGB}{255,0,71}
\definecolor{my-blue}{RGB}{0,100,163}
\definecolor{my-lightblue}{RGB}{136,195,230}
\definecolor{my-happygreen}{RGB}{146,208,80}
\definecolor{tab20-2-0}{RGB}{ 31, 119, 180}
\definecolor{tab20-2-1}{RGB}{255, 127,  14}
\definecolor{tab20-2-2}{RGB}{ 44, 160,  44}
\definecolor{tab20-2-3}{RGB}{214,  39,  40}
\definecolor{tab20-2-4}{RGB}{148, 103, 189}
\definecolor{tab20-2-6}{RGB}{227, 119, 194}
\definecolor{tab20-2-9}{RGB}{ 23, 190, 207}

\usepackage{pgfplots}
\DeclareUnicodeCharacter{2212}{−}
\usepgfplotslibrary{groupplots,dateplot,ternary}
\usetikzlibrary{patterns,shapes.arrows,angles,patterns,quotes, babel, graphs, pgfplots.groupplots,calc}
\usepackage{tkz-euclide}
\tikzset{every picture/.style={/utils/exec={\footnotesize}}}
\pgfplotsset{compat=newest}
\usepackage{neuralnetwork} 

\graphicspath{{figures/},{figures/inkscape//}}

\makeatletter
\tikzset{
    database/.style={
        path picture={
            \draw (0, 1.5*\database@segmentheight) circle [x radius=\database@radius,y radius=\database@aspectratio*\database@radius];
            \draw (-\database@radius, 0.5*\database@segmentheight) arc [start angle=180,end angle=360,x radius=\database@radius, y radius=\database@aspectratio*\database@radius];
            \draw (-\database@radius,-0.5*\database@segmentheight) arc [start angle=180,end angle=360,x radius=\database@radius, y radius=\database@aspectratio*\database@radius];
            \draw (-\database@radius,1.5*\database@segmentheight) -- ++(0,-3*\database@segmentheight) arc [start angle=180,end angle=360,x radius=\database@radius, y radius=\database@aspectratio*\database@radius] -- ++(0,3*\database@segmentheight);
        },
        minimum width=2*\database@radius + \pgflinewidth,
        minimum height=3*\database@segmentheight + 2*\database@aspectratio*\database@radius + \pgflinewidth,
    },
    database segment height/.store in=\database@segmentheight,
    database radius/.store in=\database@radius,
    database aspect ratio/.store in=\database@aspectratio,
    database segment height=0.1cm,
    database radius=0.25cm,
    database aspect ratio=0.35,
}
\makeatother %

\usepackage{catchfile}       

\usepackage[disable,colorinlistoftodos,prependcaption,linecolor=gray,backgroundcolor=gray!25,bordercolor=black]{todonotes} 

\newcommand{\todoi}[1]{\ifdefined\enabletodoi \todo[inline]{#1} \fi}
\newcommand{\tofix}[1]{\ifdefined\enabletofix \todo[inline,bordercolor=red,backgroundcolor=red!25]{\textit{Fix}: #1} \fi}

\newcommand{\toquestion}[1]{\ifdefined\enabletoquestion \todo[inline,bordercolor=cyan,backgroundcolor=cyan!25]{\textit{Q\&A}: #1} \fi}

\newcommand{\feedback}[2]{\ifdefined\enablefeedback \todo[bordercolor=lime,backgroundcolor=lime!25]{\textit{Feedback #1}: #2} \fi}

\newcommand{\norm}[1]{\lVert#1\rVert_2}

\newcommand{\first}[1]{\textbf{#1}} %
\newcommand{\snd}[1]{\underline{#1}}
\newcommand{\third}[1]{\textit{#1}}

\newcommand{\px}[2]{$#1\times#2$\,pixels}
\newcommand{\x}[3]{$#1\times#2$\,#3}

\newlength{\rightside}
\newcommand*{\rightterm}{}
\newcommand*{\term}[1]{$\displaystyle#1$}

\algnewcommand\algorithmicforeach{\textbf{for each}}
\algdef{S}[FOR]{ForEach}[1]{\algorithmicforeach\ #1\ \algorithmicdo}
 
\usepackage[
	pdfpagemode = UseNone,							%
	pdfpagelayout = TwoColumnRight,			%
	pdfauthor = {meyer},								%
	pdftitle = {yolino},							%
   colorlinks=true,         %
   hidelinks,               %
   urlcolor=blue,    %
   filecolor=blue,  %
   linkcolor=blue,  %
   citecolor=magenta,  %
   raiselinks=true,			 %
   breaklinks=true,              %
   hyperindex=true,         %
   linktoc=all,        %
   hyperfootnotes=true,     %
   bookmarks=true,          %
   bookmarksopenlevel=1,    %
   bookmarksopen=true,      %
   bookmarksnumbered=true,  %
   bookmarkstype=toc,       %
   bookmarksdepth=3,        %
   plainpages=false,        %
   pageanchor=true,         %
   pdfdisplaydoctitle=true, %
   pdfstartview=FitH,       %
   pdfpagemode=UseOutlines, %
   pdfpagelabels=true,           %
   pdfpagelayout=SinglePage, %
  ]{hyperref}

\begin{document}

\begin{acronym}[LeakyReLUXX]
\setlength{\itemsep}{0.0585cm}
\begingroup

\acro{1D}{border coordinates}

\acro{Acc}{accuracy}

\acro{CNN}{convolutional neural network}
\acroplural{CNN}[CNNs]{convolutional neural network}

\acro{DBSCAN}{Density Based Spatial Clustering of Applications with Noise}

\acro{E2E}{end-to-end}

\acro{Eu}{Euler angle}

\acro{F1}{F1 score}

\acro{ffNN}{feed-forward neural network}

\acro{FCN}{fully convolutional neural network}

\acro{FN}{false negative estimate}
\acro{FP}{false positive estimate}

\acro{FPN}{feature pyramid network}

\acro{GNN}{graph-based neural network}
\acroplural{GNN}[GNNs]{graph-based neural networks}

\acro{GT}{Ground Truth}

\acro{IoU}{intersection over union}

\acro{IPM}{inverse perspective mapping}

\acro{Kf}[Cf]{confidence deviation}
\acro{KfTP}[CfTP]{confidence deviation of TPs}

\acro{KAI}{map change dataset of Pauls et al.}

\acro{KIT}{Karlsruhe Institute of Technology}

\acro{kNN}{k-nearest neighbors}

\acro{Li}{linear activation function}

\acro{LHD}{line segment Hausdorff distance}

\acro{LReLu}[LeakyReLU]{leaky rectified linear unit}

\acro{MAE}{average absolute deviation}

\acro{MCMC}{Markov Chain Monte Carlo}

\acro{MLP}{multi-layer perceptron}

\acro{MP}[M]{midpoint coordinates}

\acro{MPNN}{message passing neural network}

\acro{MR}[MD]{midpoint direction coordinates}

\acro{MRT}{Institute of Measurement and Control}

\acro{MSE}{mean squared error}
\acroplural{MSE}[MSEs]{mean squared errors}

\acro{mussp}[MuSSP]{minimum-update successive shortest path}

\acro{NN}{neural network}
\acroplural{NN}[NNs]{neural networks} %

\acro{NMS}{non-maximum suppression}

\acro{Prec}[Pr]{precision}

\acro{Dir}[D]{direction coordinates}

\acro{Recall}[Re]{recall}

\acro{ReLu}[ReLU]{rectified linear unit}

\acro{RNN}{recurrent neural network}
\acroplural{RNN}[RNNs]{recurrent neural networks}

\acro{RMSE}{root mean squared error}

\acro{Cart}[SE]{Cartesian points}

\acro{Si}{sigmoid function}

\acro{SSE}{sum squared error}

\acro{TN}{true negative estimate}
\acro{TP}{true positive estimate}

\endgroup
\end{acronym}

\title{YOLinO++: Single-Shot Estimation of Generic Polylines for Mapless Automated Diving}

\author{Annika Meyer and Christoph Stiller,~\IEEEmembership{Fellow, IEEE}
\thanks{Annika Meyer and Christoph Stiller are with the Institute for Measurement and Control Systems at the Karlsruhe Institute of Technology (KIT), Karlsruhe, Germany. E-mail: {\ttfamily research@amyr.de}}
\thanks{}
}

\maketitle

\begin{abstract}
    In automated driving, highly accurate maps are commonly used to support and complement perception. 
    These maps are costly to create and quickly become outdated as the traffic world is permanently changing. 
    In order to support or replace the map of an automated system with detections from sensor data, a perception module must be able to detect the map features.
    We propose a neural network that follows the one shot philosophy of YOLO but is designed for detection of 1D structures in images, such as lane boundaries.
    We extend previous ideas by a midpoint based line representation and anchor definitions. 
    This representation can be used to describe lane borders, markings, but also implicit features such as centerlines of lanes.
    The broad applicability of the approach is shown with the detection performance on lane centerlines, lane borders as well as the markings both on highways and in urban areas. 
    Versatile lane boundaries are detected and can be inherently classified as dashed or solid lines, curb, road boundaries, or implicit delimitation.
\end{abstract}
\tofix{No more than 250 words.}

\section{introduction}\label{sec:introduction}

\IEEEPARstart{C}{urrent} perception algorithms mainly represent the environment as areas, points or bounding boxes. 
In automated driving, important elements, however, like lane boundaries and markings, are line-based entities. 
Thus, they are not represented ideally. 
Additionally, existing approaches for lane estimation mainly focus on highways, so intersection geometries often are not even considered.

\begin{figure}[t]
	\centering
	\includegraphics[width=1\linewidth,trim=0 448 0 128,clip]{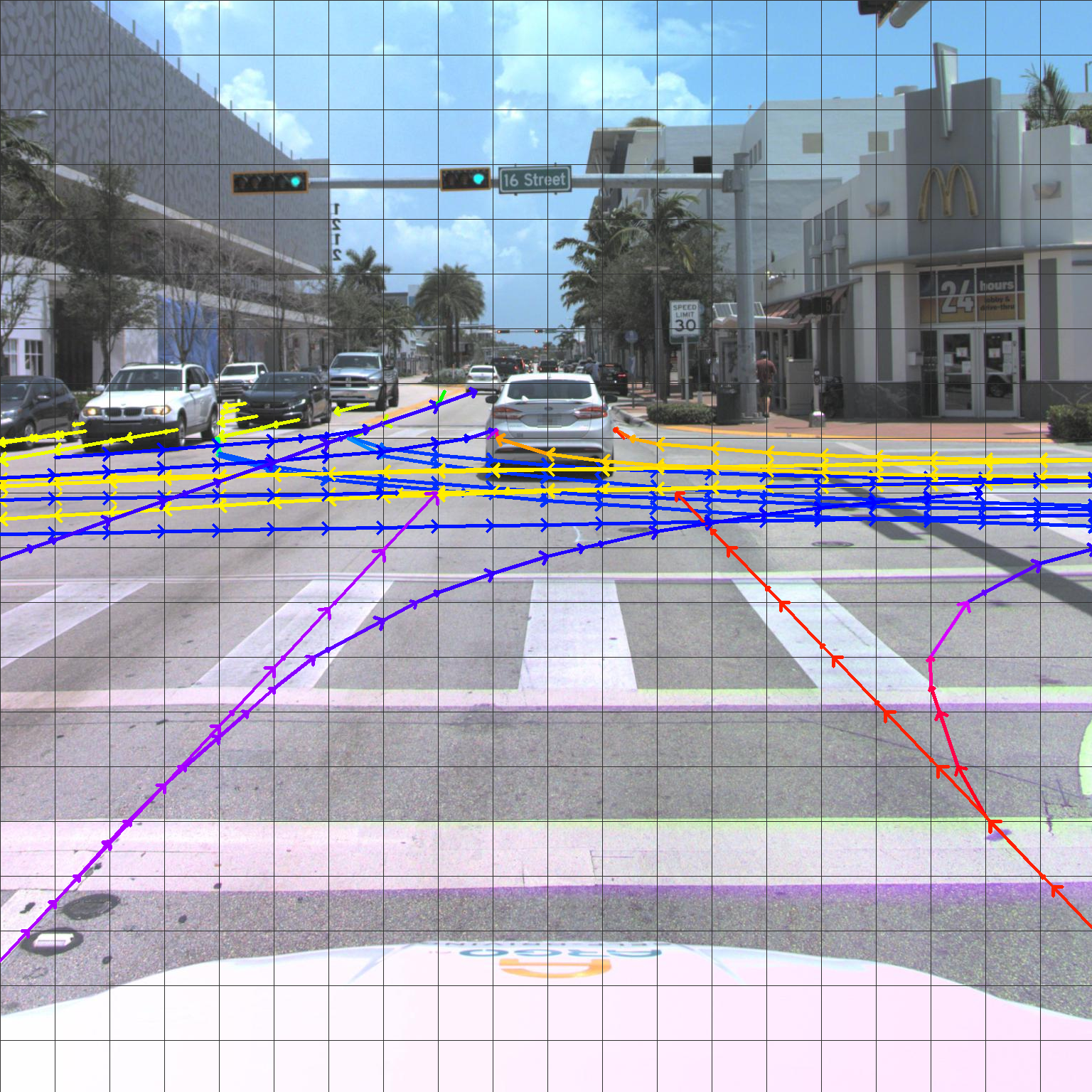}
	\caption{Estimating polylines discretized by a fixed grid enables various topologies especially for urban areas and fast estimation. We show an example image with its map projected into an image from the Argoverse dataset~\cite{wilson_ArgoverseNextGeneration_2021}.}
	\label{fig:horizon_duplicate}
\end{figure}

This work presents a method for lane estimation that combines the generalization capability of neural networks with a suitable line representation for urban space. 
Not only line entities like lane borders, but even any line-like elements can be detected and classified in real-time. 
Thus, for example, lane borders and center lines can be detected simultaneously without the need for a separate algorithm. 
This enables to perceive even complex intersection models in real time (see~\autoref{fig:horizon_duplicate}). 
Hence, the contributions are the following:
\begin{enumerate} 
    \item Proposal of a novel midpoint and direction based representation for grid-based line estimation,
	\item Ablation study on anchor-based line segment prediction in grid cells,
	\item Analysis of line segment detection with manually and data-driven anchors
    \item Extensive evaluation and ablation study on large public data sets
\end{enumerate}

The remainder of this work is organized as follows. The next section discusses related literature. Section~\ref{cha:yolino} introduces the architecture, our representation of line segments and different anchor and training strategies. Section~\ref{cha:eval} presents evaluation results on public data sets and an ablation study before we close with concluding remarks in \autoref{cha:conclusions_and_outlook}.
\section{Related Work}\label{cha:related_work}

\feedback{Stiller}{In der related works section gibst du für meinen Geschmack zu wenig Credit an andere. Du zitierst zu oftknapp und ohne den damaligen Fortschritt zu würdigen.}

In the field of automated driving, approaches have already been presented to detect line-shaped features -- particularly map features -- using learned methods. 
Most relevant are within the area of lane detection, which has evolved into dense estimation methods 
\cite{
    oliveira_EfficientDeepModels_2016,
    barnes_FindYourOwn_2017,
    meyer2018semantic,
    pan_SpatialDeepSpatial_2018,
    chen_ProgressiveLiDARAdaptation_2019,
    ghafoorian_ELGANEmbeddingLoss_2019,
    hou_LearningLightweightLane_2019,
    philion_FastDrawAddressingLong_2019,
    ort_MapLiteOnlineHD_2022
}, which classify pixels as lane, lane boundary or non-lane classes,
and sparse parametric estimation methods 
\cite{
    suleymanov_InferringRoadBoundaries_2018,
    garnett_3DLaneNetEndtoend3D_2019,
    homayounfar_DAGMapperLearningMap_2019,    
    liang_ConvolutionalRecurrentNetwork_2019,
    qin_UltraFastStructureaware_2020,
    tabelini_KeepYourEyes_2021,
    liu_CondLaneNetTopToDownLane_2021,
    zurn_LaneGraphEstimation_2021,
    liu_VectorMapNetEndtoendVectorized_2022,    
    li_HDMapNetOnlineHD_2022
    }. 

Regardless of the applied algorithm, authors either rely on images from a front camera \cite{
    oliveira_EfficientDeepModels_2016,
    barnes_FindYourOwn_2017,
    pan_SpatialDeepSpatial_2018,
    suleymanov_InferringRoadBoundaries_2018,
    chen_ProgressiveLiDARAdaptation_2019,
    ghafoorian_ELGANEmbeddingLoss_2019,
    hou_LearningLightweightLane_2019,
    philion_FastDrawAddressingLong_2019,
    garnett_3DLaneNetEndtoend3D_2019,
    qin_UltraFastStructureaware_2020,
    tabelini_KeepYourEyes_2021,
    liu_CondLaneNetTopToDownLane_2021
}, a 360 degree imagery \cite{
    li_HDMapNetOnlineHD_2022,
    liu_VectorMapNetEndtoendVectorized_2022
} or image sequences over multiple time steps \cite{
    liang_ConvolutionalRecurrentNetwork_2019,
    zurn_LaneGraphEstimation_2021,   
    ort_MapLiteOnlineHD_2022,
}. 
Since map materials are commonly represented in a top view perspective, some authors
\cite{ 
    liang_ConvolutionalRecurrentNetwork_2019,
    zurn_LaneGraphEstimation_2021,
    ort_MapLiteOnlineHD_2022
} introduce a projection of the image data into a birds eye view.

In some works, LIDAR data is also used.
The pointclouds are usually mapped into a topview grid structure with intensity \cite{
    homayounfar_DAGMapperLearningMap_2019,
    liang_ConvolutionalRecurrentNetwork_2019,
    zurn_LaneGraphEstimation_2021,
    ort_MapLiteOnlineHD_2022
} and height \cite{ 
    liang_ConvolutionalRecurrentNetwork_2019,
    zurn_LaneGraphEstimation_2021,
    ort_MapLiteOnlineHD_2022
} per cell.
HDMapNet~\cite{li_HDMapNetOnlineHD_2022} and
processes the point clouds of the LIDAR directly with a suitable architecture and learns to transform the data into a birds eye view. 
In \cite{
    liang_ConvolutionalRecurrentNetwork_2019,
    zurn_LaneGraphEstimation_2021,
    ort_MapLiteOnlineHD_2022,
    li_HDMapNetOnlineHD_2022,
    liu_VectorMapNetEndtoendVectorized_2022
} measurements of LIDAR and camera are fused.

All ideas above about the use of fused or projected data can be applied to arbitrary algorithms, because especially neural networks can be constructed modular using the encoder-decoder structure.
Thus, the encoder can be extended by the necessary fusion and transformation for each approach, so that the encoder has a low relevance concerning the consideration of suitable representation concepts.

\paragraph{Feed-Forward Neural Networks}
In order to estimate lanes, many approaches estimate the area of the ego lane~\cite{
    oliveira_EfficientDeepModels_2016}, all lanes~\cite{meyer2018semantic}
or the road area~\cite{
    oliveira_EfficientDeepModels_2016,
    chen_ProgressiveLiDARAdaptation_2019,
    fan_SNERoadSegIncorporatingSurface_2020
} with semantic segmentation.
Alternatively, it is applied to pixels that fall on the lane border (as opposed to the lane area)~\cite{
    pan_SpatialDeepSpatial_2018,
    suleymanov_InferringRoadBoundaries_2018,
    ghafoorian_ELGANEmbeddingLoss_2019,
    hou_LearningLightweightLane_2019,
    philion_FastDrawAddressingLong_2019,
    liu_CondLaneNetTopToDownLane_2021,
    li_HDMapNetOnlineHD_2022
    }. 
 
Although this point-wise or pixel-wise estimation is widely used to determine areas of the same class and already show quite useful results, the networks themselves do not infer any area or instance information in the process, but consider each pixel separately. 
Thus, they also do not explicitly learn the areas \cite{ghafoorian_ELGANEmbeddingLoss_2019}. 
This circumstance can give rise to predictions that contradict simple geometric assumptions of a lane. 

All approaches that present a one-dimensional classification system do not provide a suitable representation for intersections. 
Here, a hierarchical class structure is needed that can distinguish intersecting lanes. 

Another dense description of a lane geometry is the pixel-wise distance transform. 
Here, each pixel describes the distance \cite{
    ort_MapLiteOnlineHD_2022
    }
or the distance vector \cite{
    liang_ConvolutionalRecurrentNetwork_2019,
    homayounfar_DAGMapperLearningMap_2019} 
to the nearest lane border. 
This form of representation facilitates post-processing by providing geometric cues for the lane borders, but is ambiguous in intersecting scenarios, so that, for example, \cite{ort_MapLiteOnlineHD_2022} explicitly exclude intersections. 
 
In the TuSimple benchmark for highway scenarios, approaches predict lanes in the front image as the position of pixels in specific image rows~\cite{qin_UltraFastStructureaware_2020, liu_CondLaneNetTopToDownLane_2021,zheng_CLRNetCrossLayer_2022a}. 
With the assumption that lanes run vertically in the image, contiguous polylines are constructed from these points. 
This assumption does not hold for urban areas, since crossing arms must be described by lines running horizontally. 

A further development of the image row principle are anchors. 
Here, fixed anchor lines described along the direction of travel are defined in 2D-~\cite{suleymanov_InferringRoadBoundaries_2018,tabelini_KeepYourEyes_2021,li_LineCNNEndtoEndTraffic_2020} or 3D-space~\cite{garnett_3DLaneNetEndtoend3D_2019} and the orthogonal deviations from those are estimated. 
On the one hand, anchors introduce a simple model assumption that holds for highways and enables simple architecture for parametric estimations of e.g. lane borders. 
On the other hand, anchors defined along the direction of travel, or vertically in the image, likewise do not allow crossing lanes at intersections to be represented.
This is due to the fact that only one support point is described for each position along the anchors.
The introduction of assigning multiple predictions to an anchor \cite{garnett_3DLaneNetEndtoend3D_2019} allows for the splitting and merging of lanes. 
However, horizontal lines are still not representable.

\paragraph{Iterative Neural Networks}
In addition to \acsp{ffNN}, recurrent architectures are applied~\cite{
    homayounfar_DAGMapperLearningMap_2019,
    liang_ConvolutionalRecurrentNetwork_2019}, for example, by iteratively predicting the vertices of a polyline.
\cite{homayounfar_DAGMapperLearningMap_2019} proposes a regression on the vector to the next control point, while~\cite{liang_ConvolutionalRecurrentNetwork_2019} applies semantic segmentation to individual image segments to estimate the position of the next control point.
However, the iterative structure of these approaches lead to non-constant execution time, vanishing gradients in the training process, and large inference times.
A direction of travel of the estimated polyline is not implemented by any of the approaches mentioned so far.

Given the structure of polylines as concatenation of line segments, a \ac{GNN} can be applied to the scene graph describing line segments as nodes. 
Zürn et al.~\cite{zurn_LaneGraphEstimation_2021} predict box hypotheses with a region-proposal approach whose center points represent the support points of the polylines. 
Then, the \ac{GNN} classifies the edges between those nodes as connected or not. 
The use of rectangular hypotheses does not represent the linear structure appropriately, which cannot cope with complex lane geometries, like at intersections.

The VectorMapNet~\cite{liu_VectorMapNetEndtoendVectorized_2022} proposes a transformer architecture.
In a two-step procedure, a rough description of the polyline is first determined using two to four key points, and then a polyline is estimated for each of these descriptions.
The coordinates of the polyline are generated autoregressively, so that the estimation cannot take place in parallel, thus leading to higher execution times. 
They achieve promising results feasible for map representation. 

\paragraph{Application to Intersections}
\label{sec:rw_intersections} 

Research in lane detection often considers only non-intersecting areas of the road and ignores the complexity of intersections. 
To determine intersection geometries, it is important that an algorithm can represent merging lanes and provide a valid estimate despite occlusions. 
However, many approaches are not even capable of representing lines with arbitrary direction. 

Of the work mentioned so far that directly processes sensor data with a neural network, only \cite{
    liang_ConvolutionalRecurrentNetwork_2019,
    li_HDMapNetOnlineHD_2022,
    liu_VectorMapNetEndtoendVectorized_2022,
    zurn_LaneGraphEstimation_2021} 
present the application on a complete intersection. 
They do so using elaborate processing with an iterative network architecture~\cite{liang_ConvolutionalRecurrentNetwork_2019,
homayounfar_DAGMapperLearningMap_2019,
zurn_LaneGraphEstimation_2021}, or simple post-processing with elaborate intermediate representations~\cite{li_HDMapNetOnlineHD_2022}. 
Liang et al.~\cite{liang_ConvolutionalRecurrentNetwork_2019} also do not estimate lanes, but road borders.

In the literature, a model-based estimation is often applied to intersections estimation like in \cite{geiger_3d_2014,meyer2019mcmc}. 
So far, these methods offer promising approaches to fuse heterogeneous detections, but do not provide satisfactory results in real time yet.
Previous approaches use mostly point-based measurements, due to the lack of line-based detectors. 
This lack of measurements can be fixed by the predictions of this work. 

\paragraph{Conclusion on Related Work}
Feed-forward neural networks are not ideally suited for the structure of complex intersection models. 
In particular, the fact that intersections have a varying number of arms or lanes and thus cannot be simply described by a fixed number of variables and thus the fixed tensor of a feed-forward network.

The YOLO approaches~\cite{yolov1_redmon_you_2016,yolov2_Redmon_YOLO9000betterfaster_2017,Simony_ComplexYOLOEulerRegionProposalRealtime_2018}, that this work is inspired by, represent a set of architectures that detect a dynamic number of objects in images -- without an elaborate multistep or iterative procedure.
Instead, a single-shot \acl{ffNN} is presented in which the input image is overlaid with a grid structure. 
For each of these grid cells the network estimates hypotheses for an object detection.

Thus, they present a simple network architecture that infers results in real time with forward linked layers, converges easily, and requires no special training strategies. 
Moreover, these methods provide constant computational complexity regardless of the actual number of objects in the scene.

With the grid-based representation, extended objects can be mapped well. 
A polyline, however, does not describe an area. 
A bounding box of a polyline describing, for example, a turning lane, easily covers the entire image and is thus not very informative for the actual course.

Instead, we extend our previously presented idea called \textit{YOLinO}~\cite{meyer_YolinoGenericSingle_2021}, that adapts the ideas of \textit{YOLO} to the area of \textit{line} estimation. 
We discretize the polylines and employ the advantages of the grid:
The polylines are subdivided by the cell grid and thus predicted discretized as line segments rather than as a whole object. 
We extend this idea by a novel line representation and anchor based predictions.

\feedback{Stiller}{Am Ende von Sec2 muss *unbedingt* noch mal kommen, wo sich Yolino++ einordnet. Ich würde erst [27] erwähnen und dann erläutern, dass Yolino++ diese (und möglichst andere) aufgreift und erweitert, um folgende Dinge: ... Gibt dir die Chnce nochmal auf ein paar Highlights hinzuweien.}

\section{Spatially Discretized Line Estimation}
\label{cha:yolino}

We sample the polylines through the cell edges, thus, each resulting line segment automatically bounds itself to the cell. 
The entire polyline is then constructed from these pieces, which can flexibly represent intersections and complex topologies. 
In contrast to related work, this designs an architecture that allows for 2D discretization of the scene while requiring a very simple architecture. 

The network must be able to represent multiple possibly crossing line segments in a cell.
The concept of hypothesis estimation presented in~\cite{yolov1_redmon_you_2016} can be exploited for this purpose. 
Each cell estimates not only one line segment, but proposes multiple possibly crossing or overlapping line hypotheses.    

In contrast to~\cite{meyer_YolinoGenericSingle_2021}, the a priori distribution of the line segments is modeled by \textit{anchors}, relative to which the predictors output a deviation or a correction.
The anchors also allow the predictors to specialize, for example, on various e.g. horizontal or vertical line segments.

However, in a training example, more than one actual correct \ac{GT} line segment may well exist in a cell, so a loss function must explicitly respect this n-to-m mapping between predictors and \ac{GT} lines. 
In the spirit of responsibility that Redmon et al. have already formulated for object detection~\cite{yolov1_redmon_you_2016}, each predictor should specialize on a specific line type during training. 
Specifically, this means that, for example, a predictor specialized in horizontal segments will be activated by a high confidence only if horizontal lines are present in the scene. 
For this purpose, it is proposed to assign GT line segments to predictors based on an Euclidean distance. The details are shown in \autoref{sec:predictors}.

To finally select the correct hypotheses from the large number of predicted hypotheses, a \acf{NMS} can be appended as presented in~\cite{meyer_YolinoGenericSingle_2021}. 
This filters the hypotheses so that only valid ones remain. 

Since lanes and especially their centerlines offer the most versatility, the conceptual design in this paper focuses on lane centerlines.  We will show results on applications like markings, curbs and lane borders.

\subsection{Architecture}
\label{sec:yolino_architecture}

For the experiments, the darknet-19 architecture~\cite{yolov1_redmon_you_2016} is applied, but can be exchanged at will. 
The encoder compresses the incoming image by a factor of 32, then, the front end calculates a \x{n}{m}{grid} with $\gls{num_pred}$ predictors each describing a hypothesis. 
The number of predictors per cell $\gls{num_pred}$ thus determines the maximum number of line segments per spatial cell that the network can estimate. 

Since the prediction becomes less accurate when discretized into spatial cells, the resolution of the final cell structure is particularly important. 
In this work, we therefore investigate different scaling levels leading to a resolution of \x{8}{8}, \x{16}{16} and \px{32}{32} per cell.

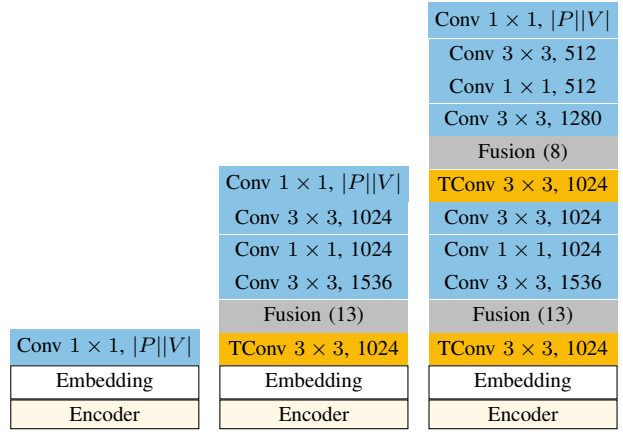
\begin{figure}[tb]
    \begin{subfigure}[b]{0.3\linewidth}
        \centering
       
\begin{tikzpicture}

\pgfmathsetmacro{\w}{2.5cm}
\pgfmathsetmacro{\d}{0}
\pgfmathsetmacro{\s}{0.3cm}

\node[rectangle,minimum width=\w,minimum height=\s,fill=my-yellow!10,draw=black] (encoder) at (0,0) {Encoder};
\node[rectangle,minimum width=\w,minimum height=\s,draw=black,above=\d of encoder.north] (embed) {Embedding};

\node[rectangle,minimum width=\w,minimum height=\s,fill=my-lightblue,align=center,above=\d of embed] (yoloconv) {Conv $1 \times 1$, $\gls{num_pred} \gls{num_vars}$};

\end{tikzpicture}
         \caption{\px{32}{32}}
    \end{subfigure}
    \begin{subfigure}[b]{0.3\linewidth}
        \centering
       
\begin{tikzpicture}

\pgfmathsetmacro{\w}{2.5cm}
\pgfmathsetmacro{\d}{0}
\pgfmathsetmacro{\s}{0.3cm}

\node[rectangle,minimum width=\w,minimum height=\s,fill=my-yellow!10,draw=black] (encoder) at (0,0) {Encoder};
\node[rectangle,minimum width=\w,minimum height=\s,draw=black,above=\d of encoder.north] (embed) {Embedding};

\node[rectangle,minimum width=\w,minimum height=\s,fill=my-yellow,align=center,above=\d of embed] (tconv1) {TConv $3 \times 3$, 1024};

\node[rectangle,minimum width=\w,minimum height=\s,fill=lightgray,align=center,above=\d of tconv1] (fusion1) {Fusion (13)};

\node[rectangle,minimum width=\w,minimum height=\s,fill=my-lightblue,align=center,above=0 of fusion1] (conv1) {Conv $3 \times 3$, 1536};
\node[rectangle,minimum width=\w,minimum height=\s,fill=my-lightblue,align=center,above=0 of conv1] (conv2) {Conv $1 \times 1$, 1024};
\node[rectangle,minimum width=\w,minimum height=\s,fill=my-lightblue,align=center,above=0 of conv2] (conv3) {Conv $3 \times 3$, 1024};

\node[rectangle,minimum width=\w,minimum height=\s,fill=my-lightblue,align=center,above=\d of conv3] (yoloconv) {Conv $1 \times 1$, $\gls{num_pred} \gls{num_vars}$};

\end{tikzpicture}         \caption{\px{16}{16}}
    \end{subfigure}
    \begin{subfigure}[b]{0.3\linewidth}
        \centering
       
\begin{tikzpicture}

\pgfmathsetmacro{\w}{2.5cm}
\pgfmathsetmacro{\d}{0}
\pgfmathsetmacro{\s}{0.3cm}

\node[rectangle,minimum width=\w,minimum height=\s,fill=my-yellow!10,draw=black] (encoder) at (0,0) {Encoder};
\node[rectangle,minimum width=\w,minimum height=\s,draw=black,above=\d of encoder.north] (embed) {Embedding};

\node[rectangle,minimum width=\w,minimum height=\s,fill=my-yellow,align=center,above=\d of embed] (tconv1) {TConv $3 \times 3$, 1024};

\node[rectangle,minimum width=\w,minimum height=\s,fill=lightgray,align=center,above=\d of tconv1] (fusion1) {Fusion (13)};

\node[rectangle,minimum width=\w,minimum height=\s,fill=my-lightblue,align=center,above=0 of fusion1] (conv1) {Conv $3 \times 3$, 1536};
\node[rectangle,minimum width=\w,minimum height=\s,fill=my-lightblue,align=center,above=0 of conv1] (conv2) {Conv $1 \times 1$, 1024};
\node[rectangle,minimum width=\w,minimum height=\s,fill=my-lightblue,align=center,above=0 of conv2] (conv3) {Conv $3 \times 3$, 1024};

\node[rectangle,minimum width=\w,minimum height=\s,fill=my-yellow,align=center,above=\d of conv3] (tconv2) {TConv $3 \times 3$, 1024};

\node[rectangle,minimum width=\w,minimum height=\s,fill=lightgray,align=center,above=\d of tconv2] (fusion2) {Fusion (8)};

\node[rectangle,minimum width=\w,minimum height=\s,fill=my-lightblue,align=center,above=0 of fusion2] (conv4) {Conv $3 \times 3$, 1280};
\node[rectangle,minimum width=\w,minimum height=\s,fill=my-lightblue,align=center,above=0 of conv4] (conv5) {Conv $1 \times 1$, 512};
\node[rectangle,minimum width=\w,minimum height=\s,fill=my-lightblue,align=center,above=0 of conv5] (conv6) {Conv $3 \times 3$, 512};

\node[rectangle,minimum width=\w,minimum height=\s,fill=my-lightblue,align=center,above=\d of conv6] (yoloconv) {Conv $1 \times 1$, $\gls{num_pred} \gls{num_vars}$};

\end{tikzpicture} 
         \caption{\px{8}{8}}
    \end{subfigure}
	\caption{Three decoder variants with different resolution per cell. Folded layers are visualized in blue, transposed folded layers in orange, and fusion layers in gray.     
    The fusion layers additionally get information from the skip connections passed on by the feature maps of the eighth or 13th folding layer from the encoder.}
	\label{fig:decoder}
\end{figure}

To estimate line segments in higher resolution the architecture is extended with additional (transposed) convolutional layers (see \autoref{fig:decoder}) so that cell structures with \x{8}{8} and \px{16}{16} cells can be predicted. 

In order to not lose precision due to compression on embedding, we employ skip connections between layers of the same size in the encoder and decoder.
As with the YOLO architectures, \acs{LReLu} is used as the activation function of the hidden layers.

\subsection{Line Representation}
\label{sec:lines}

\toquestion{English does not have start points but rather two end points in one line.}

Each estimated hypothesis describes a line segment as $\gls{segment_grid} = (\gls{pred_geom}, \gls{pred_class}, \gls{pred_conf})$. It consists of a geometric description $\gls{pred_geom}$, an optional classification $\gls{pred_class}$, and a confidence $\gls{pred_conf}$.
The confidence is used to threshold valid predictions after successful training. 

\tofix{start/end point with a gap}
\tofix{center / midpoint confusion}
\begin{figure}[tb]
	\centering
	\footnotesize
	\begin{subfigure}[b]{0.4\linewidth}
		\centering
		\def\svgwidth{\linewidth}
\begingroup%
  \makeatletter%
  \providecommand\color[2][]{%
    \errmessage{(Inkscape) Color is used for the text in Inkscape, but the package 'color.sty' is not loaded}%
    \renewcommand\color[2][]{}%
  }%
  \providecommand\transparent[1]{%
    \errmessage{(Inkscape) Transparency is used (non-zero) for the text in Inkscape, but the package 'transparent.sty' is not loaded}%
    \renewcommand\transparent[1]{}%
  }%
  \providecommand\rotatebox[2]{#2}%
  \newcommand*\fsize{\dimexpr\f@size pt\relax}%
  \newcommand*\lineheight[1]{\fontsize{\fsize}{#1\fsize}\selectfont}%
  \ifx\svgwidth\undefined%
    \setlength{\unitlength}{595.27559055bp}%
    \ifx\svgscale\undefined%
      \relax%
    \else%
      \setlength{\unitlength}{\unitlength * \real{\svgscale}}%
    \fi%
  \else%
    \setlength{\unitlength}{\svgwidth}%
  \fi%
  \global\let\svgwidth\undefined%
  \global\let\svgscale\undefined%
  \makeatother%
  \begin{picture}(1,1)%
    \lineheight{1}%
    \setlength\tabcolsep{0pt}%
    \put(0,0){\includegraphics[width=\unitlength,page=1]{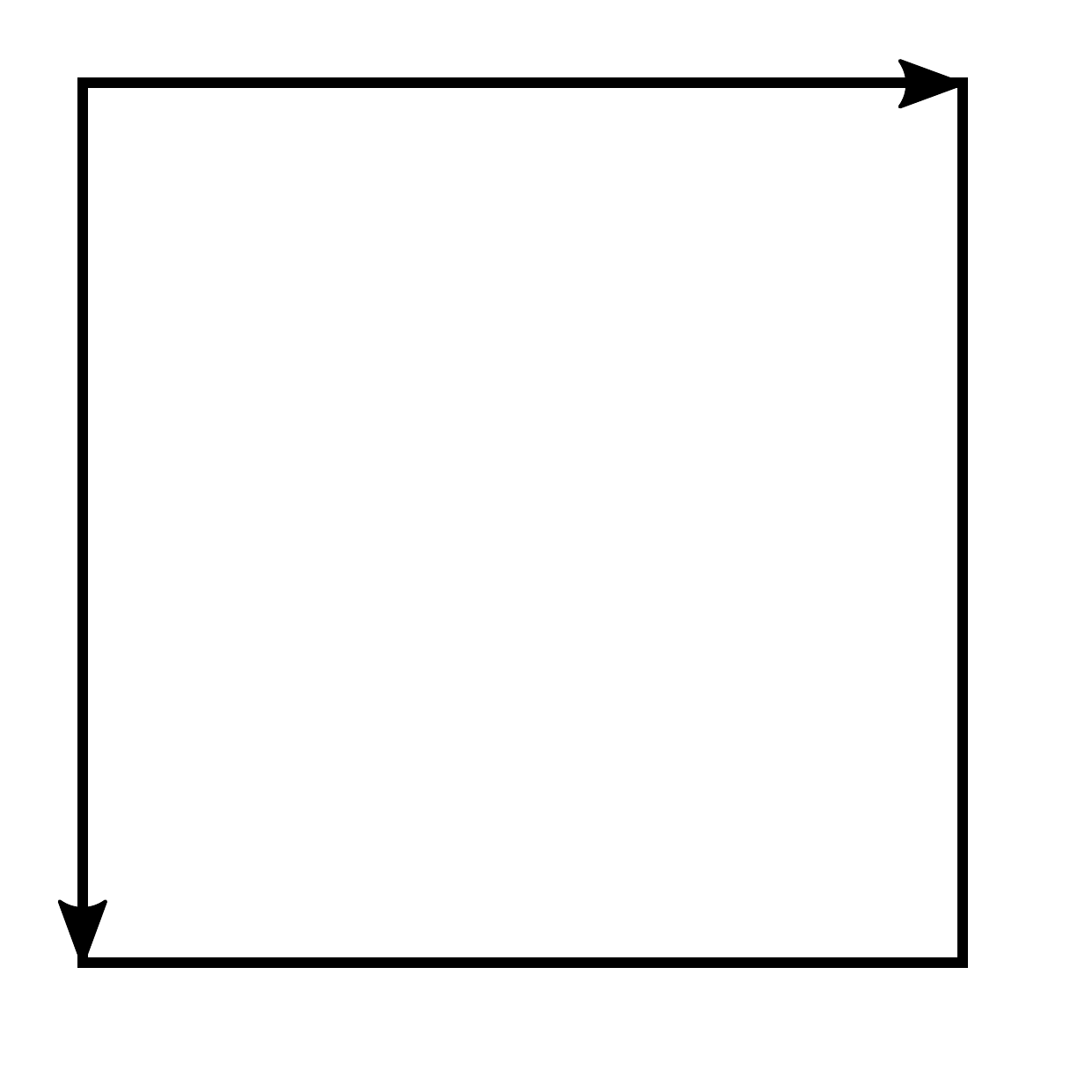}}%
    \put(0,0.94960318){\color[rgb]{0,0,0}\makebox(0,0)[lt]{\lineheight{1.25}\smash{\begin{tabular}[t]{l}0/0\end{tabular}}}}%
    \put(0.02519841,0.04246034){\color[rgb]{0,0,0}\makebox(0,0)[lt]{\lineheight{1.25}\smash{\begin{tabular}[t]{l}1\end{tabular}}}}%
    \put(0.90714286,0.94960318){\color[rgb]{0,0,0}\makebox(0,0)[lt]{\lineheight{1.25}\smash{\begin{tabular}[t]{l}1\end{tabular}}}}%
    \put(0.75595238,0.68501983){\color[rgb]{0,0,0}\makebox(0,0)[lt]{\lineheight{1.25}\smash{\begin{tabular}[t]{l}$\gls{pred_end}$\end{tabular}}}}%
    \put(0.13859127,0.1558532){\color[rgb]{0,0,0}\makebox(0,0)[lt]{\lineheight{1.25}\smash{\begin{tabular}[t]{l}$\gls{pred_start}$\end{tabular}}}}%
    \put(0,0){\includegraphics[width=\unitlength,page=2]{linerep_po.pdf}}%
  \end{picture}%
\endgroup%
 		\caption{\acs{Cart}}
		\label{fig:points}
	\end{subfigure}
	\begin{subfigure}[b]{0.4\linewidth}
		\centering
		\def\svgwidth{\linewidth}
\begingroup%
  \makeatletter%
  \providecommand\color[2][]{%
    \errmessage{(Inkscape) Color is used for the text in Inkscape, but the package 'color.sty' is not loaded}%
    \renewcommand\color[2][]{}%
  }%
  \providecommand\transparent[1]{%
    \errmessage{(Inkscape) Transparency is used (non-zero) for the text in Inkscape, but the package 'transparent.sty' is not loaded}%
    \renewcommand\transparent[1]{}%
  }%
  \providecommand\rotatebox[2]{#2}%
  \newcommand*\fsize{\dimexpr\f@size pt\relax}%
  \newcommand*\lineheight[1]{\fontsize{\fsize}{#1\fsize}\selectfont}%
  \ifx\svgwidth\undefined%
    \setlength{\unitlength}{595.27559055bp}%
    \ifx\svgscale\undefined%
      \relax%
    \else%
      \setlength{\unitlength}{\unitlength * \real{\svgscale}}%
    \fi%
  \else%
    \setlength{\unitlength}{\svgwidth}%
  \fi%
  \global\let\svgwidth\undefined%
  \global\let\svgscale\undefined%
  \makeatother%
  \begin{picture}(1,1)%
    \lineheight{1}%
    \setlength\tabcolsep{0pt}%
    \put(0,0){\includegraphics[width=\unitlength,page=1]{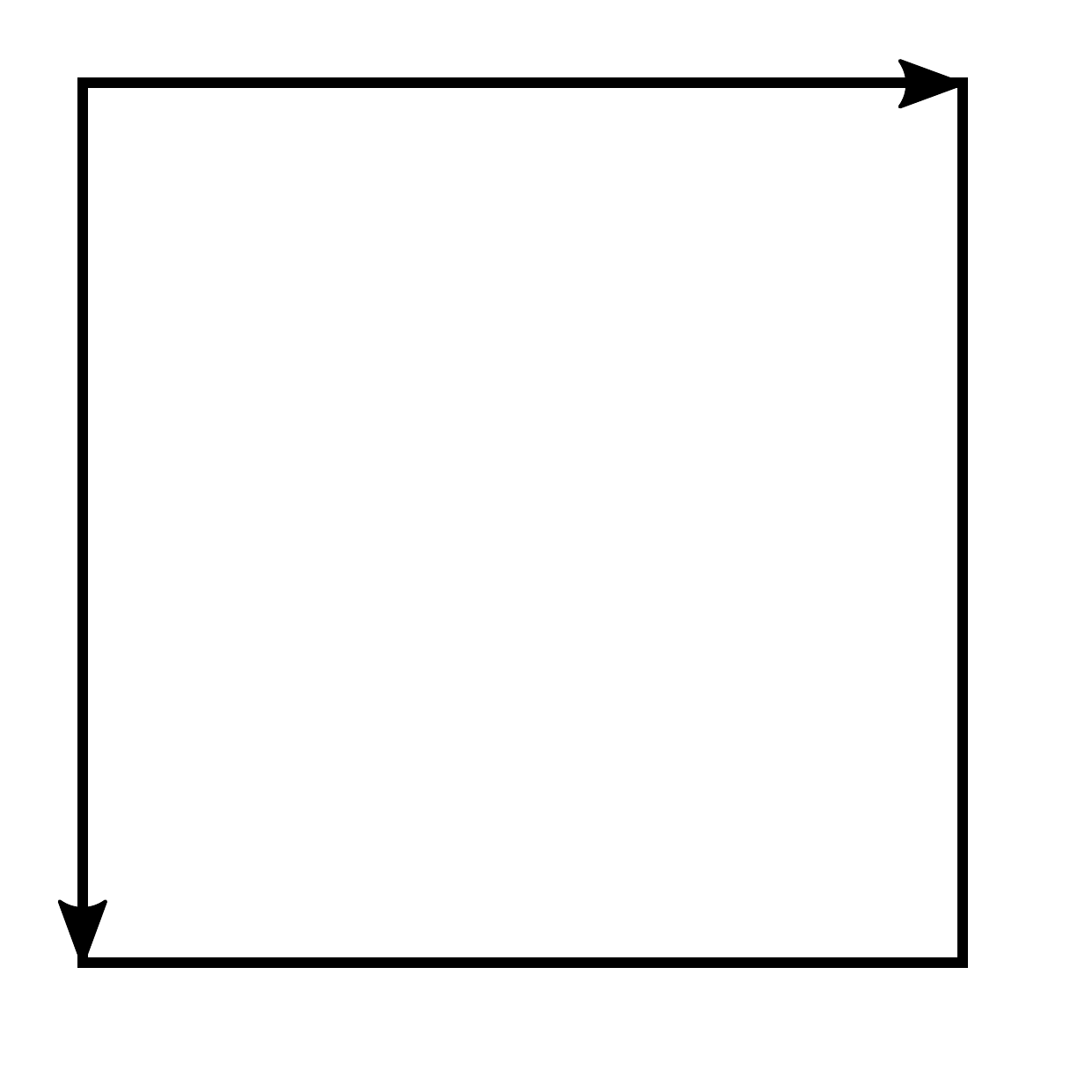}}%
    \put(0,0.94960318){\color[rgb]{0,0,0}\makebox(0,0)[lt]{\lineheight{1.25}\smash{\begin{tabular}[t]{l}0/0\end{tabular}}}}%
    \put(0.02519841,0.04246034){\color[rgb]{0,0,0}\makebox(0,0)[lt]{\lineheight{1.25}\smash{\begin{tabular}[t]{l}1\end{tabular}}}}%
    \put(0.90714286,0.94960318){\color[rgb]{0,0,0}\makebox(0,0)[lt]{\lineheight{1.25}\smash{\begin{tabular}[t]{l}1\end{tabular}}}}%
    \put(0.67405752,0.50863095){\color[rgb]{0,0,0}\makebox(0,0)[lt]{\lineheight{1.25}\smash{\begin{tabular}[t]{l}$\gls{pred_dir}_x$\end{tabular}}}}%
    \put(0.37797619,0.19995059){\color[rgb]{0,0,0}\makebox(0,0)[lt]{\lineheight{1.25}\smash{\begin{tabular}[t]{l}$\gls{pred_dir}_y$\end{tabular}}}}%
    \put(0.30238095,0.62202381){\color[rgb]{0,0,0}\makebox(0,0)[lt]{\lineheight{1.25}\smash{\begin{tabular}[t]{l}$\gls{pred_midp}$\end{tabular}}}}%
    \put(0,0){\includegraphics[width=\unitlength,page=2]{linerep_md.pdf}}%
  \end{picture}%
\endgroup%
 		\caption{\acs{MR}}
		\label{fig:md}
	\end{subfigure}
	\caption{\acf{Cart} and \acf{MR} to represent a line segment within a cell.}
	\label{fig:linerep}
\end{figure}

\subsubsection{Cartesian Points}   
As presented in~\cite{meyer_YolinoGenericSingle_2021} line segments can be described as by \acf{Cart} (see \autoref{fig:points}). 
It defines a line segment as an ordered set of two points $\gls{segment_grid}_{\gls{geom_cart}} := (\gls{pred_start},\gls{pred_end})$ with $\gls{pred_start},\gls{pred_end} \in [0,1]^2$.
The origin of the coordinate system is located in the upper left corner of each cell. 
For uniformity, the coordinate system is normalized to the cell such that the cell width (and height) equal exactly 1.

\subsubsection{Midpoint and Direction}
Since \ac{Cart} penalizes deviations along the line in the same way as parallel displacements, we propose an alternative representation that considers position and rotation in a decoupled way.
In \acf{MR} (see \autoref{fig:md}), each line segment is defined as $\gls{segment_grid}_{\gls{geom_md}} := (\gls{pred_midp},\gls{pred_dir})$ with $\gls{pred_midp} \in [0,1]^2, \gls{pred_dir} \in [-1,1]^2$. 
Here, $\gls{pred_midp}$ represents the center point and $\gls{pred_dir}$ represents the x and y portions of the vector, respectively. 

In this representation, it is important to note that the orientation is described by the absolute $x$ or $y$ component.
Here, alternatively the cosine/sine part of the orientation and an additional length parameter could be used. 
However, this leads on the one hand to the fact that five instead of four parameters must be optimized. 
On the other hand, the inference of the exact cosine/sine components depends on each other, because only selected combinations satisfy $\cos^2{\alpha} + \sin^2{\alpha} = 1$. 
Ideally, however, the target formulations should be independently learnable. 
We have confirmed this theoretical conjecture in experiments: With a cosine/sine/length representation, the networks converge with difficulty if at all.

\subsection{Predictors and Anchors}
\label{sec:predictors}
 
Ideally, a line segment is always predicted by the same predictor, as this allows specialized feature extractors to form as well as ensuring a stable training result. 
For this purpose, a clear assignment of the line segments of the \ac{GT} to a predictor has to be done first. 
Especially in intersection scenarios (see \autoref{fig:horizon_duplicate}) several line segments have to be estimated per cell, which intersect or run in parallel. 
Likewise, images offer the challenge that very similar line segments fall into the same cell close to the horizon.

As the number of predictors per cell increases, a predictor is responsible for fewer and fewer line segments in the data set, so the number directly determines how much the network must generalize for a predictor or how specialized it can be. 
The more predictors the architecture provides, the more fine-grained line types can be distinguished. 
At the same time, the influence of each training example on one predictor is reduced.
So it is important to balance the number of predictors and find a good trade-off.

In this work, on the one hand, a \textit{dynamic assignment}~\cite{meyer_YolinoGenericSingle_2021} is investigated, where in each training step the assignment is recomputed based on the prediction. 
On the other hand, \textit{anchors} are retrieved from the dataset and permanently assigned to the predictors so that the \ac{GT} line segments can be assigned before the training. 

\subsubsection{Dynamic Assignment}
\label{sec:learn_anchors}
For dynamic mapping, the \ac{GT} lines are compared to the prediction after each inference in training. 
Within each cell, we calculate the Euclidean distance in the full parameter space for each pair and determine the best 1-to-1 assignment using the Hungarian method~\cite{munkres_AlgorithmsAssignmentTransportation_1957}. 

The advantage of this approach is the automatic specialization of the predictor to individual line types. 
Since all predictors are initialized identically, they are initially located in the same point of the cell. 
So the predictors specialize automatically during training with different GT lines, so that different predictor types gradually form without explicitly specifying them.

\subsubsection{Anchors}
\label{sec:anchors}

As an alternative, a-priori knowledge can be used to determine a fixed assignment. 
On the one hand, these representatives determine the geometric description of a predictor and, on the other hand, serve as geometric anchors from which only a deviation has to be predicted. 

The advantage of anchors consists on the one hand in a clearly faster training time. 
Since no mapping has to be computed during training. 
A large part of the computations is omitted compared to the dynamic mapping.
On the other hand the prediction is initialized with an a-priori estimate, as the deviation of anchors instead of an absolute position is learned.
                           
For line segments, anchors can be specified in a uniformly distributed or a data-driven manner. 
First, it is important to note that all hypotheses are estimated simultaneously, and thus each anchor can get assigned a single GT. 
Consequently, this means that each GT line only appears in the training if no other line has been assigned to the corresponding anchor.  
Thus, the choice and distribution of the anchors is of elementary importance for the training success.

\begin{figure}[tb]
	\centering
	\begin{subfigure}[b]{0.29\linewidth}
		\centering
		\includegraphics[width=\linewidth,frame]{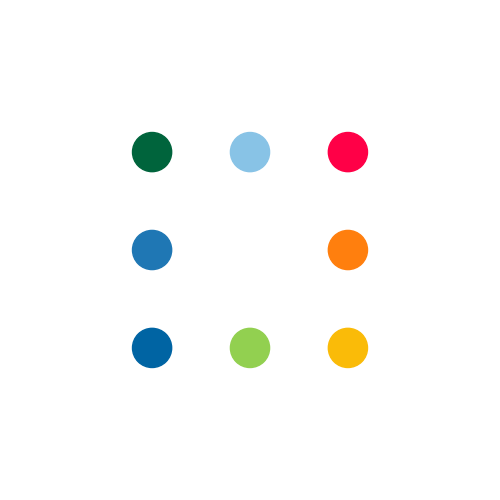}
        \label{fig:mp_manual_anchor}
	\end{subfigure}
	\begin{subfigure}[b]{0.29\linewidth}
		\centering
		\includegraphics[width=\linewidth,frame]{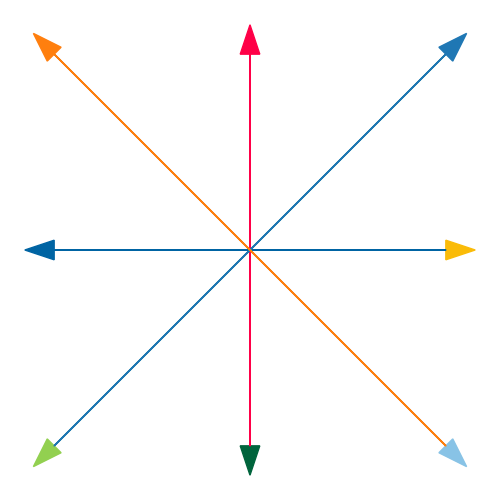}
        \label{fig:dir_manual_anchor}
	\end{subfigure}
	\begin{subfigure}[b]{0.29\linewidth}
		\centering
		\includegraphics[width=\linewidth,frame]{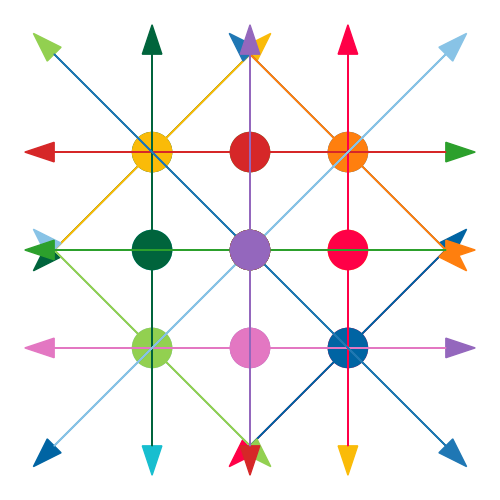}
        \label{fig:mr2_manual_anchor}
	\end{subfigure}

    \begin{subfigure}[b]{0.29\linewidth}
        \centering
        \includegraphics[width=\linewidth,frame]{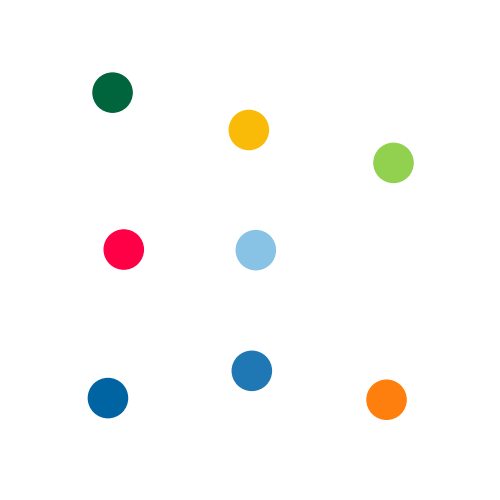}
        \caption{8 \acs{MP}}
    \end{subfigure}
    \begin{subfigure}[b]{0.29\linewidth}
        \centering
        \includegraphics[width=\linewidth,frame]{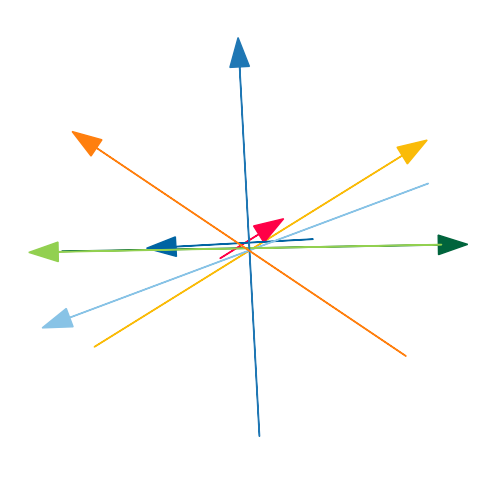}
        \caption{8 \acs{Dir}}
    \end{subfigure}
    \begin{subfigure}[b]{0.29\linewidth}
        \centering
        \includegraphics[width=\linewidth,frame]{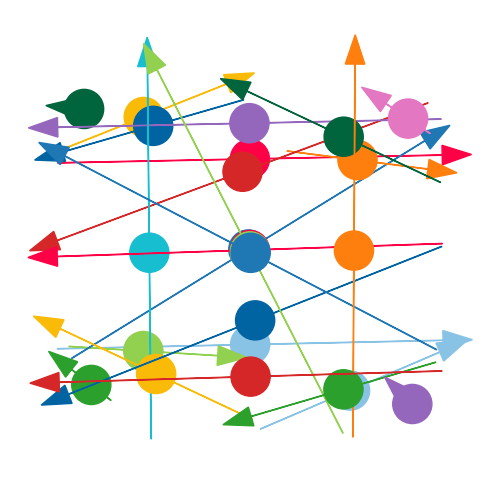}
		\caption{24 \acs{MR}} 
    \end{subfigure}
   
	\caption{Anchors for the \acs{MR} representation. In the \textit{top} row we show examples of the equally distributed anchors, whereas the \textit{bottom} row shows examples retrieved as cluster representatives from the Argoverse 2.0 dataset. The colors distinguish different anchors. Here, not all anchors are fully recognizable as they e.g. share the same center point and might travel in the opposite direction occupying the same pixel space. a) visualizes eight anchors defined and distributed by their center point. b) shows eight anchors distributed along the direction. c) shows the case for 24 anchors distributed in both center point and direction coordinates.} 
    \label{fig:anchors}
\end{figure}

\tofix{Use midpoint again!}
\toquestion{Is it possible to have a line that does not run through the midpoint?}

Anchors for centerlines, lane borders, or markings in urban areas should ideally cover the entire feature space \textit{equally}. 
Therefore, the anchors in \autoref{fig:anchors} in the top row are recommended. 

As an alternative, analogous to Redmon and Farhadi~\cite{yolov2_Redmon_YOLO9000betterfaster_2017}, \textit{data-driven anchors} can be determined from the dataset using $k$-means clustering. 
To simplify the balance between model complexity and specialization, clustering for the \acs{MR} is performed both on the midpoint~(\acs{MP}) and direction~(\acs{Dir}) as standalone cluster features, as well as on the combination~(\acs{MR}) of the two. 
For the \acl{Cart}, anchors can only be defined on the full feature space.
The cluster representatives extracted from the Argoverse dataset are shown in \autoref{fig:anchors} in the bottom row.

\subsection{Loss Function}
\label{sec:loss}
\feedback{Stiller}{Für meine Begriffe wird [27] etwas zu häufig zitiert. Das erweckt den Eindruck als gäbe der Artikel nix Neues her, weshalb man ihn ablehnen sollte. Um so wichtiger ist es, in der related work die neuen Beiträge auch ggü [27] substantiell darzulegen}

We formulate the loss for each training sample as the weighted sum of geometric loss, classification loss and confidence loss~\cite{meyer_YolinoGenericSingle_2021}.
Since there are more predictors than \ac{GT}, and a GT segment only exist in a few cells at all, the geometric and classification loss is determined only for predictors assigned to a GT. 
The confidence loss, on the other hand, is calculated for each predictor.
\autoref{eq:loss} shows the calculation of the 
individual partial loss functions for a line segment $\gls{segment_grid} = (\gls{pred_geom}, \gls{pred_class}, \gls{pred_conf})$ compared to a GT segment $\gls{gt_segment_grid} \in \gls{gt_set_segments_grid}$.

\renewcommand*{\rightterm}{\gls{weight_conf1} (\gls{pred_conf} - 1)^2, }
\settowidth{\rightside}{\term{\rightterm}}

\begin{align}
    \nonumber
	\gls{loc_loss}(\gls{segment_grid},\gls{gt_segment_grid}) &= 
        \begin{cases}
            \makebox[\rightside][l]{\term{\gls{distance}(\gls{pred_geom}, \hat{\gls{pred_geom}}), }}&\text{if } \gls{segment_grid} \text{ is assigned to } \gls{gt_segment_grid}\\
            0, &\text{else}\\
        \end{cases}\\ 
	\label{eq:loss}
	\gls{conf_loss}(\gls{segment_grid},\gls{gt_segment_grid}) &= 
        \begin{cases}
			\rightterm&\text{if } \gls{segment_grid} \text{ is assigned to } \gls{gt_segment_grid}\\
            \gls{weight_conf0}(\gls{pred_conf} - 0)^2, &\text{else}\\
        \end{cases}\\
    \nonumber
	\gls{class_loss}(\gls{segment_grid},\gls{gt_segment_grid}) &= 
        \begin{cases} 
            \makebox[\rightside][l]{\term{(\gls{pred_class} - \hat{\gls{pred_class}})^2, }}&\text{if } \gls{segment_grid} \text{ is assigned to } \gls{gt_segment_grid}\\ 
            0, &\text{else} \\
        \end{cases}
\end{align}

The \textit{geometric loss} $\gls{loc_loss}$ is determined by the Euclidean distance $\gls{distance}$.
With the \textit{confidence loss} $\gls{conf_loss}$, a high confidence $\gls{pred_conf}$ is either rewarded or penalized depending on whether a predictor is assigned to a \ac{GT}.
For the \textit{classification loss} $\gls{class_loss}$ of the predictor, Redmon and Farhadi propose~\cite{yolov2_Redmon_YOLO9000betterfaster_2017} the squared distance between the predicted pseudo probabilities $\gls{pred_class}(c)$ for each class $c \in \gls{set_classes_gt}$ and the one-hot encoding of the \ac{GT}. 
The weights $\omega$ of each loss term can be determined empirically.
Learning the loss weights as additional variables as presented in~\cite{kendall_MultitaskLearningUsing_2018} does not lead to any improvement in our experiments.

\section{Evaluation}
\label{cha:eval}

For this work, we demonstrate that different map features can be detected accurately on centerlines, markings of various types and lane borders using three datasets. 
They also provide settings with different perspectives: front view and aerial images. 

The {\textit{Argoverse}} 2.0 Dataset~\cite{wilson_ArgoverseNextGeneration_2021} includes camera and lidar data as well as a high accuracy map. 
With the lidar depth information included, the existing map material can be projected into the images and thus used as GT data. 
For this work, we use the centerlines and borders of each lane, respectively. 
The image sequences are reduced to every 20th image as successive images have little relevance for training. 
Since we are interested in the lanes, the front image is cropped to a square section of \px{1536}{1536} at the bottom of the image, which is then scaled to \px{640}{640}.
The map data within a radius of \SI{80}{\meter} around the ego-vehicle pose is projected into the front camera image using the provided software and converted into image coordinates.

The \textit{\acf{KAI}}~\cite{pauls_CanWeTrust_2018} includes annotations for aerial images of German highways. 
Here, the positions of solid and dashed markings in the aerial images were given as polylines that can be projected onto the aerial images. 

{\textit{TuSimple}}\footnote{TuSimple \url{https://github.com/TuSimple/tusimple-benchmark}} is a dataset that includes images from a front camera on highway scenes 
annotated with the lane borders of the direction of travel.

\subsection{Evaluation Metric}
\label{sec:ex_metrics}

We compare different design aspects of the approach with \textit{retrieval metrics} and the \acs{MAE} of geometric aspects. 
For each predictor in a cell, we calculate metrics between the predictor and the assigned GT. 
The assignment is done within each cell as described in \autoref{sec:predictors}.
When using anchors, the association is already given by the assignment prior to training. 
In dynamic association, the association is determined after inference by minimizing the association cost. 

For the retrieval metrics, a prediction is considered positive if the confidence $c > 0.5$. 
A prediction is termed True-positive~(\acs{TP}) if it is associated with a \ac{GT} line segment, thus a true-negative estimate~(\acs{TN}) describes a prediction with low confidence that is not assigned to a GT line segment. 
If the combination of confidence and assignment does not fit then the predictions are said to be a \acf{FP} or a \acf{FN}.
For the whole mapping, \acf{Recall}, \acf{Prec}, \acf{F1} and \acf{Acc} are calculated. 

For all true-positive line segments, we also present the \acs{MAE} regarding:
    the Cartesian points~($\norm{\cdot}$),
    the midpoints~(MP),
    the lengths~(L) and
    the confidences~(\acs{KfTP}).
In addition, the experiments consider the difference of confidences among all predictions~(\acs{Kf}).

In order to be able to compare dynamic and anchor-based assignments, we calculate all scores compared to the actual ground truth, thus, duplicate assignments are regarded within the measures. 
However, we also provide the average percentage of duplicate assignments in the dataset as (MA). 
                                                
\subsection{Training System}   
\label{sec:train_system}

\todoi{The trainings in this chapter are built on a base experiment that has already been shown to be a suitable configuration in preliminary experiments. The configuration of the base experiment is listed in \autoref{ap:base}.}

To achieve better generalization, the images are slightly augmented at each training step.
We adapt the brightness, contrast, and hue, crop and rotate the image, and fill areas randomly.
Image mirroring is not recommended for lane detection, as this reverses the direction of travel on a road and places the ego-vehicle in the left lane. 
Each training batch consists of 64 images, except for experiments with an increased grid resolution. Here, we use 32 images. 

Four experiments are run simultaneously on a node with two Intel Xeon Platinum 8368 CPUs (38~cores each) with four NVIDIA A100-40 GPUs (\SI{40}{\giga\byte}) \footnote{The trainings in this work were performed on the supercomputer HoreKa, which is funded by the Ministry of Science, Research and the Arts of Baden-Württemberg and the Federal Ministry of Education and Research}.
The implementation was done using PyTorch 1.11.0, CUDA 11.3 and Python 3.6.8 on a Red Hat 8.4 Linux system.

\subsection{Experiments}
\label{sec:experiments}

\autoref{tab:ex_all} presents the results of the trainings on the validation set of Argoverse. 
In the first block, we investigate the \textit{line representations} and \textit{activation functions} (Sigmoid \acs{Si}, linear \acs{Li}) for the geometry or the confidence, respectively.
The \acs{MR} representation achieves a better F1 measure regardless of the activation function.
Confidences are estimated more precisely with a sigmoid activation regardless of the line representation. 

\todoi{add inference time in table }

\begin{table*}[tb]
    \centering 
    \caption{Experiments comparing different configurations on the Argoverse validation set. 
    All measures are evaluated with the model at convergence. 
    The epoch of convergence is listed at Cv. 
    The average inference of the network takes \SI{3}{\milli\second} per image with a batch size of 64 images.
calculation    To improve the readability, we highlight the \first{best}, \snd{second best}, and \third{third best} results for each column.} 
    \label{tab:ex_all}
    
\begin{tabular}{|cc|cccrcr|rrrr|r|rrrr|r|r|r|}
    \hline
    &            & Line       & Anchor     & Act.  & \gls{num_pred} & Cluster & $\square$ & \acs{F1}     & \acs{Recall} & \acs{Prec}   & \acs{Acc}    & \acs{Kf}     & \acs{KfTP}   & $\norm{\cdot}$ & MP           & L            & Cv & $t_{\mathcal{L}}$ & MA\\
    \hline
    \hline
    \parbox[t]{2mm}{\multirow{4}{*}{\rotatebox[origin=c]{90}{Line}}} &                 & \acs{Cart} & \acs{Cart} & Li,Si & 8              & kM      & 32        & 0.42         & 0.45         & 0.39         & 0.96         & 0.06         & 0.22         & 3.58           & 3.70         & 2.62         & 27 & 29            & 20 \\
    &             & \acs{Cart} & \acs{Cart} & Li,Li & 8              & kM      & 32        & 0.42         & 0.45         & 0.40         & 0.96         & 0.08         & 0.23         & 3.55           & 3.69         & \third{2.53} & 48 & \third{28}               & 20 \\              \\
    & $\star$     & \acs{MR}   & \acs{MR}   & Li,Si & 8              & kM      & 32        & 0.44         & 0.46         & 0.43         & 0.96         & 0.06         & 0.16         & 3.52           & 3.05         & 4.04         & 63   & 29                  & 20 \\
    &             & \acs{MR}   & \acs{MR}   & Li,Li & 8              & kM      & 32        & {0.44}       & {0.47}       & 0.42         & 0.96         & 0.07         & 0.21         & 3.53           & 3.05         & 4.19         & 51 & \third{28}                & 20 \\
    \hline
    \hline
    \parbox[t]{2mm}{\multirow{11}{*}{\rotatebox[origin=c]{90}{Predictors \& Anchors}}} &            & \acs{MR}   & \acs{Dir}  & Li,Si & 6              & kM      & 32        & \third{0.45} & \snd{0.50}   & 0.41         & 0.95         & 0.07         & \third{0.15} & 4.22           & 3.77         & 4.60         & 48 & \snd{27}               & 21 \\
    &            & \acs{MR}   & \acs{Dir}  & Li,Si & 8              & kM      & 32        & {0.44}       & {0.47}       & {0.43}       & 0.96         & 0.05         & {0.16}       & 4.06           & 3.66         & 4.36         & 45 & \snd{27}                  & 21 \\
    &            & \acs{MR}   & \acs{Dir}  & Li,Si & 8              & gl      & 32        & \snd{0.47}   & {0.47}       & \snd{0.48}   & \third{0.97} & \third{0.04} & \first{0.13} & 5.69           & 4.76         & 6.87         & 42 & \first{26}                & 28 \\
    &            & \acs{MR}   & \acs{Dir}  & Li,Si & 24             & kM      & 32        & 0.38         & 0.37         & 0.39         & \first{0.99} & \first{0.02} & 0.22         & 3.09           & 2.98         & 3.80         & 54 & 29                        & 17 \\
    &            & \acs{MR}   & \acs{MP}   & Li,Si & 6              & kM      & 32        & 0.41         & 0.43         & 0.39         & 0.95         & 0.08         & 0.22         & 7.58           & 7.19         & 6.51         & 33 & \third{28}                & 20 \\
    &            & \acs{MR}   & \acs{MP}   & Li,Si & 8              & kM      & 32        & 0.39         & 0.41         & 0.38         & 0.96         & 0.06         & 0.21         & 6.86           & 6.62         & 5.43         & 78 & \snd{27}                  & 20 \\
    &            & \acs{MR}   & \acs{MP}   & Li,Si & 8              & gl      & 32        & 0.40         & 0.41         & 0.38         & 0.96         & 0.06         & 0.20         & 6.87           & 6.65         & 7.89         & 72 & \third{28}                & 21 \\
    &            & \acs{MR}   & \acs{MP}   & Li,Si & 24             & kM      & 32        & 0.29         & 0.29         & 0.30         & \snd{0.98}   & \snd{0.03}   & 0.30         & 6.78           & 6.80         & 5.19         & 69 & 30                        & 13 \\

    &             & \acs{MR}   & \acs{MR}   & Li,Si & 6              & kM      & 32        & {0.44}       & \third{0.48} & 0.41         & 0.95         & 0.07         & 0.19         & 3.83           & 3.27         & 4.28         & 30 & \snd{27}                 & 20 \\
    &             & \acs{MR}   & \acs{MR}   & Li,Si & 24             & kM      & 32        & 0.38         & 0.41         & 0.36         & \first{0.99} & \snd{0.03}   & 0.21         & \third{2.32}   & \third{2.00} & 2.62         & 57 & 30                       & 15 \\
    &             & \acs{MR}   & \acs{MR}   & Li,Si & 24             & gl      & 32        & 0.39         & 0.41         & 0.38         & \first{0.99} & \first{0.02} & 0.20         & 2.83           & 2.36         & 3.34         & 57 & 29                       & 16 \\
    \hline
    \hline
    \parbox[t]{2mm}{\multirow{3}{*}{\rotatebox[origin=c]{90}{Scale}}} &       & \acs{MR}   & \acs{MR}   & Li,Si & 8              & kM      & 32        & \third{0.45} & 0.46         & \third{0.44} & 0.96         & 0.06         & {0.16}       & 3.47           & 3.02         & 4.01         & 75 & 35*               & \multicolumn{1}{|r|}{-} \\
    & $\oplus$    & \acs{MR}   & \acs{MR}   & Li,Si & 8              & kM      & 16        & 0.38         & 0.41         & 0.35         & \snd{0.98}   & 0.05         & 0.25         & \snd{1.62}     & \snd{1.44}   & \snd{1.92}   & 30 & 76*               & \multicolumn{1}{|r|}{-} \\
    & $\diamond$  & \acs{MR}   & \acs{MR}   & Li,Si & 8              & kM      & 8         & 0.28         & 0.40         & 0.22         & \snd{0.98}   & \third{0.04} & 0.28         & \first{0.78}   & \first{0.71} & \first{0.97} & 27 & 151*              & \multicolumn{1}{|r|}{-} \\
    \hline
    \hline
    Dyn. &$\sim$   & \acs{MR}   & -          & Li,Si & 8              & none    & 32        & \first{0.61} & \first{0.64} & \first{0.59} & \third{0.97} & 0.05         & \snd{0.14}   & 4.75           & 4.01         & 4.58         & 51 & 197               & 0 \\
    \hline
\end{tabular}
\end{table*}  

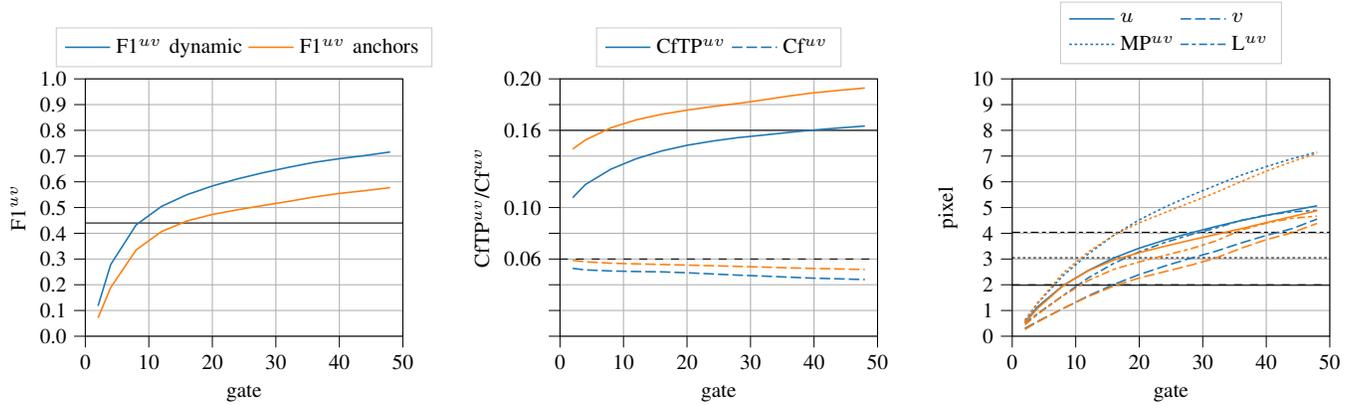
\begin{figure*}[h]
    \centering
    \begin{subfigure}[b]{0.32\textwidth}
       
\begin{tikzpicture}

\definecolor{color0}{rgb}{0.12156862745098,0.466666666666667,0.705882352941177}
\definecolor{color1}{rgb}{1,0.498039215686275,0.0549019607843137}

\begin{axis}[
legend cell align={left},
legend columns=2,
legend style={
  fill opacity=0.8,
  draw opacity=1,
  text opacity=1,
  at={(0.5,1.2)},
  anchor=north,
  draw=white!80!black
},
grid,
tick align=outside,
tick pos=left,
width=\linewidth,
x grid style={white!69.0196078431373!black},
xlabel={gate},
xmin=0, xmax=50,
xtick style={color=black},
xtick={-10,0,10,20,30,40,50,60},
xticklabels={\ensuremath{-}10,0,10,20,30,40,50,60},
y grid style={white!69.0196078431373!black},
ylabel={F1$^{uv}$},
ymin=0, ymax=1,
ytick style={color=black},
ytick={0,0.1,0.2,0.3,0.4,0.5,0.6,0.7,0.8,0.9,1.0},
yticklabels={0.0,0.1,0.2,0.3,0.4,0.5,0.6,0.7,0.8,0.9,1.0},
]

\draw [black] (0,0.44) -- (50,0.44); 

\addplot [semithick, color0]
table {%
2 0.117550238005261
4 0.278739674678805
8 0.432571770390334
12 0.504607279861603
16 0.549890643412619
20 0.583866284330855
24 0.611575064875306
28 0.635315988625191
32 0.656295541812976
36 0.67550858515599
40 0.689822034519605
44 0.70201086146838
48 0.715935498462901
};
\addlegendentry{F1$^{uv}$ dynamic};
\addplot [semithick, color1]
table {%
2 0.0712633912474172
4 0.189177192799103
8 0.335296639873813
12 0.406821828763983
16 0.447796889864231
20 0.473110739456701
24 0.491397466988975
28 0.508120153381958
32 0.523904662853356
36 0.540944761347906
40 0.554814879278684
44 0.565584402154076
48 0.577413809972236
};
\addlegendentry{F1$^{uv}$ anchors};

\end{axis}

\end{tikzpicture}
        
    \end{subfigure}
    \hfill
    \begin{subfigure}[b]{0.32\textwidth}
       
\begin{tikzpicture}

\definecolor{color0}{rgb}{0.12156862745098,0.466666666666667,0.705882352941177}
\definecolor{color1}{rgb}{1,0.498039215686275,0.0549019607843137}

\begin{axis}[
legend cell align={left},
legend columns=2,
legend style={
  fill opacity=0.8,
  draw opacity=1,
  text opacity=1,
  at={(0.5,1.2)},
  anchor=north,
  draw=white!80!black
},
grid,
tick align=outside,
tick pos=left,
width=\linewidth,
x grid style={white!69.0196078431373!black},
xlabel={gate},
xmin=0, xmax=50,
xtick style={color=black},
xtick={-10,0,10,20,30,40,50,60},
xticklabels={\ensuremath{-}10,0,10,20,30,40,50,60},
y grid style={white!69.0196078431373!black},
ylabel={CfTP$^{uv}$/Cf$^{uv}$},
ymin=0, ymax=0.2,
ytick style={color=black},
ytick={0.0,0.02,0.04,0.06,0.08,0.1,0.12,0.14,0.16,0.18,0.2},
yticklabels={ , , ,0.06, ,0.10, , ,0.16, ,0.20},
]

\draw [black] (0,0.16) -- (50,0.16); 
\draw [black,dashed] (0,0.06) -- (50,0.06); 

\addplot [semithick, color0]
table {%
2 0.107706360921666
4 0.117874475749763
8 0.129794581113635
12 0.137921540519676
16 0.143959231876038
20 0.14837655606302
24 0.151640208790431
28 0.154310277990393
32 0.156290958056579
36 0.158381310266417
40 0.160262276997437
44 0.16185957918296
48 0.163366293182244
};
\addlegendentry{CfTP$^{uv}$}
\addplot [semithick, color0, dash pattern=on 4pt off 1.5pt]
table {%
2 0.0527077272937104
4 0.0515477433800697
8 0.0507193712165226
12 0.0502878476840418
16 0.0500059930255284
20 0.0493116766415737
24 0.0483929788744127
28 0.0474398468394537
32 0.0467338134106752
36 0.0458985223762086
40 0.0450937036122824
44 0.0445825053832015
48 0.0439982070914796
};
\addlegendentry{Cf$^{uv}$}
\addplot [semithick, color1, forget plot]
table {%
2 0.145454864042836
4 0.15251801988563
8 0.161937753090987
12 0.168144173718788
16 0.172466971584268
20 0.175687467326989
24 0.17845807365469
28 0.180887225914646
32 0.183632370587942
36 0.186654878226486
40 0.189196748507989
44 0.191095757726076
48 0.192895704024547
};
\addplot [semithick, color1, dash pattern=on 4pt off 1.5pt, forget plot]
table {%
2 0.0590122972388525
4 0.0578317227395805
8 0.0567492243607302
12 0.0562210455536842
16 0.055755368157013
20 0.0552687501987895
24 0.0547941722781271
28 0.0542541263071266
32 0.0537248866984973
36 0.0531345547453777
40 0.0526513829827308
44 0.0522696420147612
48 0.0518461140627796
};
\end{axis}

\end{tikzpicture}
        
    \end{subfigure}
    \hfill
    \begin{subfigure}[b]{0.32\textwidth} 
       
\begin{tikzpicture}

\definecolor{color0}{rgb}{0.12156862745098,0.466666666666667,0.705882352941177}
\definecolor{color1}{rgb}{1,0.498039215686275,0.0549019607843137}

\begin{axis}[
legend cell align={left},
legend columns=2,
legend style={
  fill opacity=0.8,
  draw opacity=1,
  text opacity=1,
  at={(0.5,1.3)},
  anchor=north,
  draw=white!80!black
},
grid,
tick align=outside,
tick pos=left,
width=\linewidth,
x grid style={white!69.0196078431373!black},
xlabel={gate},
xmin=0, xmax=50,
xtick style={color=black},
xtick={-10,0,10,20,30,40,50,60},
xticklabels={\ensuremath{-}10,0,10,20,30,40,50,60},
y grid style={white!69.0196078431373!black},
ylabel={pixel},
ymin=0, ymax=10,
ytick style={color=black},
ytick={-1,0,1,2,3,4,5,6,7,8,9,10},
yticklabels={\ensuremath{-}1,0,1,2,3,4,5,6,7,8,9,10}
]

\draw [black] (0,1.98) -- (50,1.98); 
\draw [dashed,black] (0,2) -- (50,2); 
\draw [dash pattern=on 1pt off 1pt,black] (0,3.05) -- (50,3.05); 
\draw [dash pattern=on 3pt off 1.25pt on 1.5pt off 1.25pt,black] (0,4.04) -- (50,4.04); 

\addplot [semithick, color0]
table {%
2 0.550972305439614
4 1.14276133679055
8 1.97053996936695
12 2.55741044637319
16 3.04113855233064
20 3.42179402789554
24 3.7348286783373
28 4.007322826901
32 4.25856548386651
36 4.49231955811784
40 4.69366146422721
44 4.8838520501111
48 5.06861336166794
};
\addlegendentry{$u$}
\addplot [semithick, color0, dash pattern=on 4pt off 1.5pt]
table {%
2 0.304685083595482
4 0.575256928398802
8 1.07374526197846
12 1.57001190249984
16 2.01047577084722
20 2.39606193271843
24 2.72554062508248
28 3.02455954938321
32 3.296853033272
36 3.62020085953377
40 3.91633155539229
44 4.17460203170776
48 4.55128324998392
};
\addlegendentry{$v$};
\addplot [semithick, color0, dash pattern=on 1pt off 1pt]
table {%
2 0.620645563344698
4 1.28449538591746
8 2.35409235954285
12 3.19955314172281
16 3.92623891701569
20 4.52275291649071
24 5.01445217390318
28 5.45349639170879
32 5.85773578849999
36 6.25380976135666
40 6.58992597219106
44 6.8935975126318
48 7.15486575461723
};
\addlegendentry{MP$^{uv}$};
\addplot [semithick, color0, dash pattern=on 3pt off 1.25pt on 1.5pt off 1.25pt]
table {%
2 0.502103984355927
4 0.878411446068738
8 1.57900325994234
12 2.27120723273303
16 2.83238959956814
20 3.28023504566502
24 3.63843395258929
28 3.93849397994377
32 4.19991836676726
36 4.4965376016256
40 4.7058228286537
44 4.83322749266753
48 4.8890365523261
};
\addlegendentry{L$^{uv}$};
\addplot [semithick, color1, forget plot]
table {%
2 0.515519486085789
4 1.08345358758359
8 1.95573605395652
12 2.5498859141324
16 2.96092857541265
20 3.24807621337272
24 3.49358757122143
28 3.7281109771213
32 3.95054110320839
36 4.17450491802112
40 4.41350925290907
44 4.64551129856625
48 4.87778704875224
};
\addplot [semithick, color1, dash pattern=on 4pt off 1.5pt, forget plot]
table {%
2 0.263105377152159
4 0.550641436834593
8 1.08211842420939
12 1.54404345396403
16 1.94912763866218
20 2.26251902773574
24 2.50819819037979
28 2.75323341666041
32 3.0325405339937
36 3.41501535596074
40 3.75156987679971
44 4.01757333085344
48 4.38295079566337
};
\addplot [semithick, color1, dash pattern=on 1pt off 1pt, forget plot]
table {%
2 0.589409074267826
4 1.29414991752521
8 2.45354745839093
12 3.28316736865688
16 3.92571418349807
20 4.40938913499987
24 4.80049258309442
28 5.18763386236655
32 5.57573009181667
36 6.00643320341368
40 6.41156539401493
44 6.77859872096294
48 7.10232735968925
};
\addplot [semithick, color1, dash pattern=on 3pt off 1.25pt on 1.5pt off 1.25pt, forget plot]
table {%
2 0.440411719116005
4 0.847374524619128
8 1.57107103193128
12 2.16706011102006
16 2.6017542948594
20 2.88761214307837
24 3.13903629457628
28 3.38909981701825
32 3.69538013355152
36 4.1218263845186
40 4.39454930537456
44 4.57754129332465
48 4.67024017669059
};
\end{axis}

\end{tikzpicture}
        
    \end{subfigure}
    \caption{Comparing the $uv$-metrics for a dynamic assignment (blue) and anchors (orange). We evaluate the same models as in \autoref{tab:ex_all}, marked with $\sim$ and $\star$, respectively, but applied a circular evaluation gate in $uv$ coordinates for determining \acsp{TP}. The results in \autoref{tab:ex_all} are presented with a rectangular matching gate limited to the cell. The horizontal lines mark the cell based estimation for anchors.}
    \label{fig:dynamic_f1}
\end{figure*}

For the \textit{anchors} of the predictors, we need to define both the cluster strategy ($k$-means or uniform) and the cluster space. 
Since the \acs{MR} consists of \acf{MP} and \acf{Dir}, both the combination and the individual dimensions are examined. 
First, we examine how often more than one \acs{GT} line segment is assigned to the same anchor. 
Any second assignment cannot appear in the training because there is no predictor that receives the corresponding \ac{GT}-line segment as a training target(multiple assignments~(MA) given in average percentage).
Here, we find that a higher number of anchors leads to fewer duplicates. 
Additionally, duplication occurs less frequently on average in the midpoint space than in the other spaces.
When comparing the $k$-means anchors and the equally distributed anchors, the $k$-means anchors have fewer duplicates.  

Although the \acl{MP} produce fewer duplicate assignments, training with these anchors still achieves a worse F1 measure than the other models. 
Moreover, the geometric deviations are almost twice as high as for the other anchor variants.
Comparing $k$-means and uniformly distributed anchors, the {equally distributed anchors always exhibit a better F1-measure} than the $k$-means variant with the same number of predictors. 
At the same time, however, the {geometric distances, especially for direction anchors, are significantly worse} than for the $k$-means variants. 
Accordingly, it is necessary to prioritize between geometric precision and high F1-measure based on the application.

\feedback{Stiller}{Das könnte ein Hinweis auf zu wenig Training data sein, oder?}
In general, the fewer \textit{predictors} the network has available the higher the F1 measure. 
This is because with few predictors, a single predictor is significantly more often responsible and therefore more often assigned costs to learn from. 
At the same time, the geometric precision increases with the number of predictors, because a predictor can specialize more and more in certain geometric properties. 
The number of training epochs necessary for convergence increases with the number of predictors.

The number of cells and thus their \textit{resolution} can be increased with an extension of the decoder.
A batch size of 32 must be chosen for the experiments, since the variants with the largest scaling require larger memory than the previous trainings. 
The scaling of the grid shows differences especially in the inference time: with \px{32}{32} it takes \SI{10}{\milli\second}, for 16 pixels it takes \SI{27}{\milli\second} and for 8 pixels \SI{42}{\milli\second} per image in a batch of 32 images. 
\todoi{put times to table}
The precision and thus the F1 measure, decreases with each increase in the number of cells, so that the {initially chosen scale factor of 32 gives the best F1 measure}.
This is due to the noisy GT data, through which the position of the centerline cannot be determined precisely and consistently. 
This causes the GT lines to scatter across cells, especially at the bottom of the image, which is reflected as an inaccuracy in the prediction.

The geometric error of the prediction decreases at each scaling step, so that the mean error at a resolution of \px{8}{8} is in the subpixel range.
With an appropriate \acf{NMS}~\cite{meyer_YolinoGenericSingle_2021}, the number of false positives can be significantly reduced while maintaining geometric precision, so that {scaling to \px{8}{8} cells in combination with a \acs{NMS} is a reasonable option}.
However, it is important to weigh whether the {two to four times the inference time for higher scales} is justified for each application individually. 

\autoref{fig:dynamic_f1} further provides insight for comparing the dynamic assignment with the anchors across different matching gates in the evaluation. 
Here, we provide the retrieval metrics within a matching gate measured in $uv$ coordinates in the full image, compared to the evaluation in \autoref{tab:ex_all} limited by the cell borders. 
As the \textit{dynamic assignment} improves all measures it is clearly preferable to anchors. 
However, if fast training is needed, this assessment does not hold, as training with {predefined anchors converges faster and has shorter training times}.

The following recommendations can be derived from the results of the ablation studies: 
\begin{enumerate}
    \item Both \acf{MR} and Cartesian points~(\acs{Cart}) are suitable for estimation in the cell grid. 
    \acs{MR} achieves better training results. 
    \item The number of predictors must be weighed depending on the application. The more predictors there are, the better the geometry of the line segments is estimated, but the precision of the confidence estimate decreases. 
    \item In the comparison between anchor definitions, anchors determined in a data-driven manner on the full \acs{MR} cluster space using $k$-means clustering provide the best training results.
    \item Scaling the cell grid is associated with higher noise, but is recommended in conjunction with a \acs{NMS}. The improved results must be weighed against the higher inference time for an application. 
    \item Dynamic assignment during training achieves a better F1-measure than training with anchors, but exhibits significantly longer training times.  
\end{enumerate}

\subsection{Qualitative Results}
\label{sec:qual_eval}

\begin{figure*}[ht]
    \centering
    \begin{subfigure}[b]{0.32\linewidth}
        \centering
        \includegraphics[width=1\linewidth]{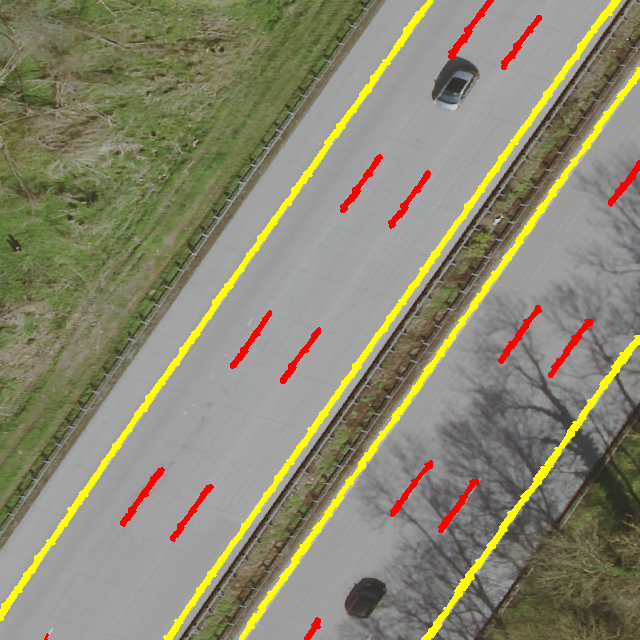}
    \end{subfigure}
    \begin{subfigure}[b]{0.32\linewidth}
        \centering
        \includegraphics[width=1\linewidth]{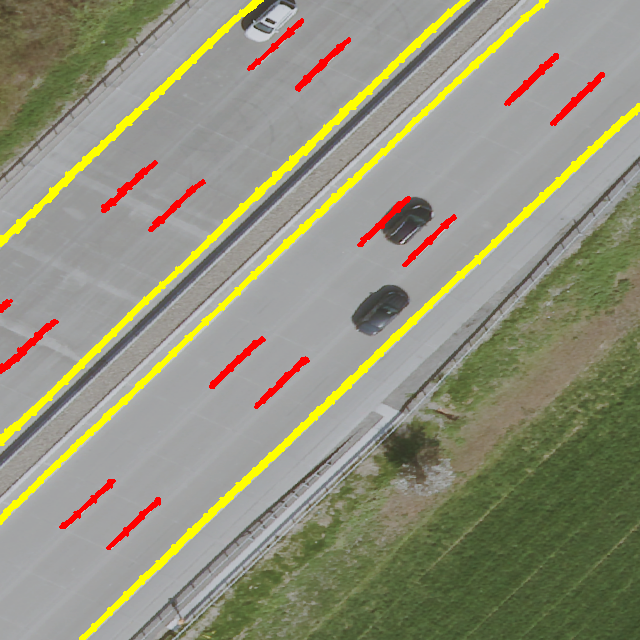}
    \end{subfigure}
    \begin{subfigure}[b]{0.32\linewidth}
        \centering
        \includegraphics[width=1\linewidth]{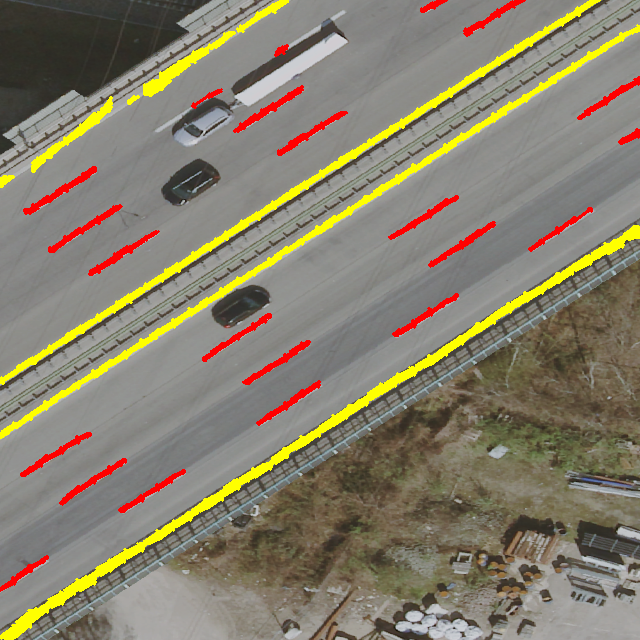}
    \end{subfigure}
    \caption{Predictions on the \acs{KAI} test dataset published in~\cite{meyer_YolinoGenericSingle_2021}. Line segment classification is used here to distinguish heterogeneous objects. All variants also estimate a direction of the marking and thus correctly describe the direction of travel of the lane. Image source of aerial photos: \textcopyright City of Karlsruhe | Liegenschaftsamt}
    \label{fig:qual_eval_aerial}
\end{figure*}

To get a qualitative sense of the results, this section presents the prediction on the TuSimple, \acs{KAI} and Argoverse dataset. 
In \autoref{fig:qual_eval_aerial}, we show example results for the application to detect markings in aerial images.
Here, the line segments describe two classes of markers: solid and dashed. 
The direction of the line segment can be used to estimate the direction of the lanes. 
The direction is thereby implicitly and continuously estimated over the order of the Cartesian points. 
As shown in~\cite{meyer_YolinoGenericSingle_2021} the estimation achieves a point-based F1 measure of \SI{89}{\percent} and has slight difficulties only with unusual objects (see right image). 
\todoi{explain point based f1}

\begin{figure*}[ht]
    \centering
    \begin{subfigure}[b]{0.49\linewidth}
        \centering
       
        \includegraphics[width=1\linewidth,trim=3 240 464 35,clip]{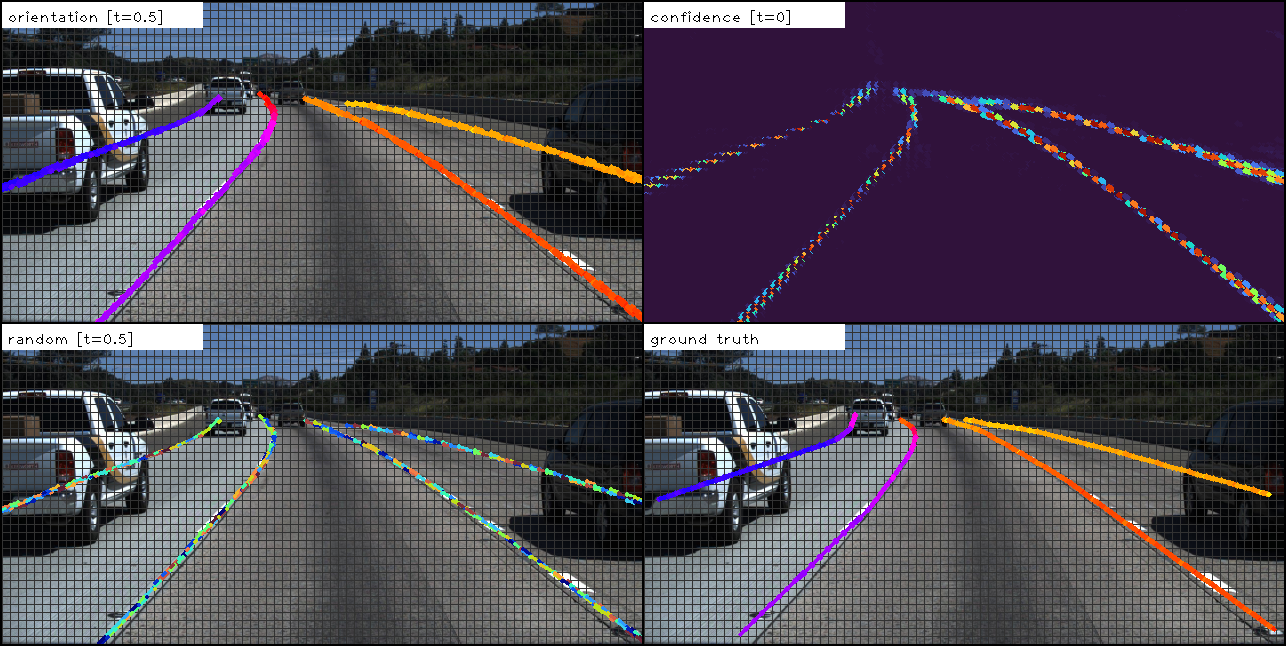}
    \end{subfigure}    
    \begin{subfigure}[b]{0.49\linewidth}
        \centering
        \includegraphics[width=1\linewidth,trim=3 240 464 35,clip]{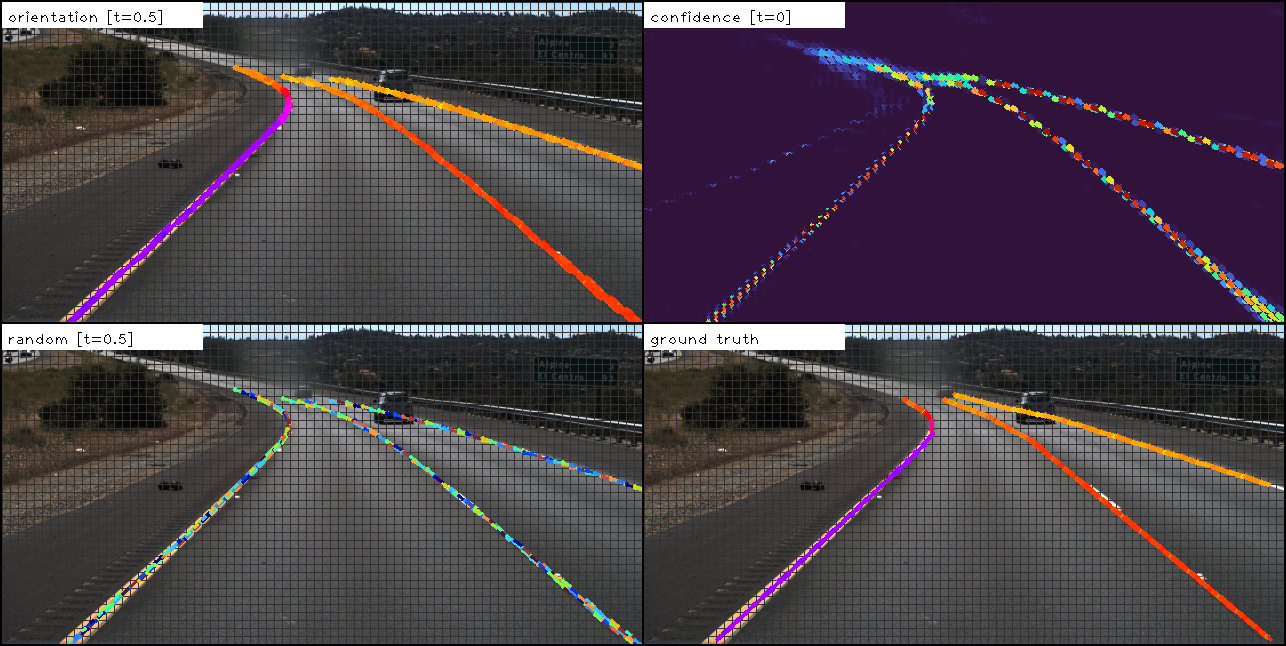}
    \end{subfigure}

    \begin{subfigure}[b]{0.49\linewidth}
        \centering
        \includegraphics[width=1\linewidth,trim=3 323 643 35,clip]{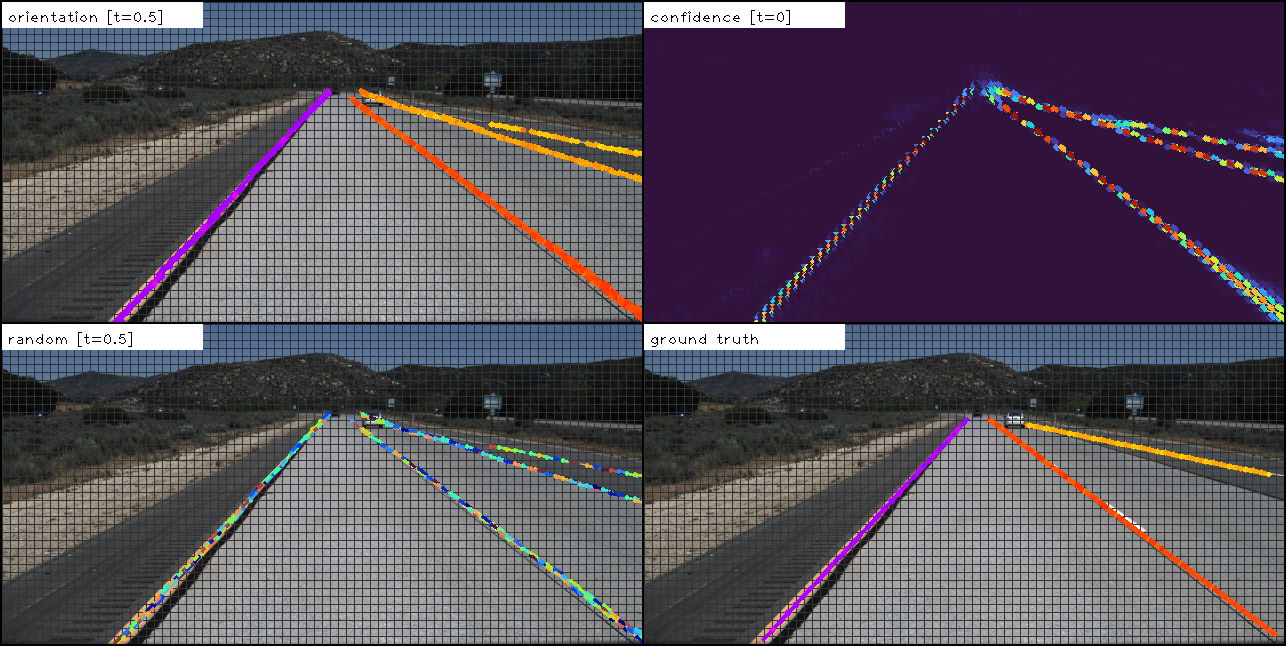}
    \end{subfigure}    
    \begin{subfigure}[b]{0.49\linewidth}
        \centering
       
        \includegraphics[width=1\linewidth,trim=3 232 464 27,clip]{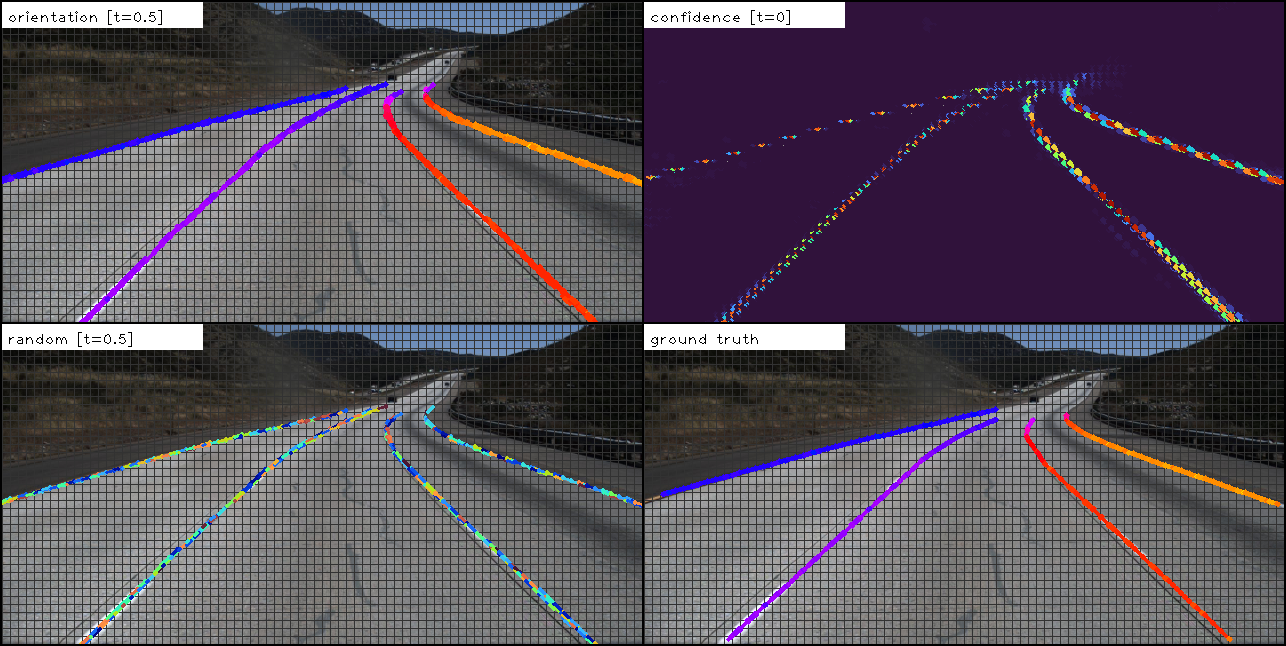}
    \end{subfigure}    
    \caption{Predictions on TuSimple dataset with \px{8}{8} per cell. Line segments are color coded according to their orientation in the image, showing the coherence of the estimated direction.}
    \label{fig:qual_eval_tus}
\end{figure*}

\autoref{fig:qual_eval_tus} shows results of different scenes on the TuSimple dataset. 
Here, we again can make use of the implicit direction encoding. 
We estimate the lanes with the direction of travel. 
As shown in \cite{meyer_YolinoGenericSingle_2021}, the results reach the level of related work, without using any model assumptions specialized for highways typically applied in related work.
\feedback{Stiller}{Tusimple Tabelle aus [27]}

\begin{figure*}
    \begin{subfigure}[b]{0.49\linewidth}
        \includegraphics[width=\linewidth,trim=2 464 464 112,clip]{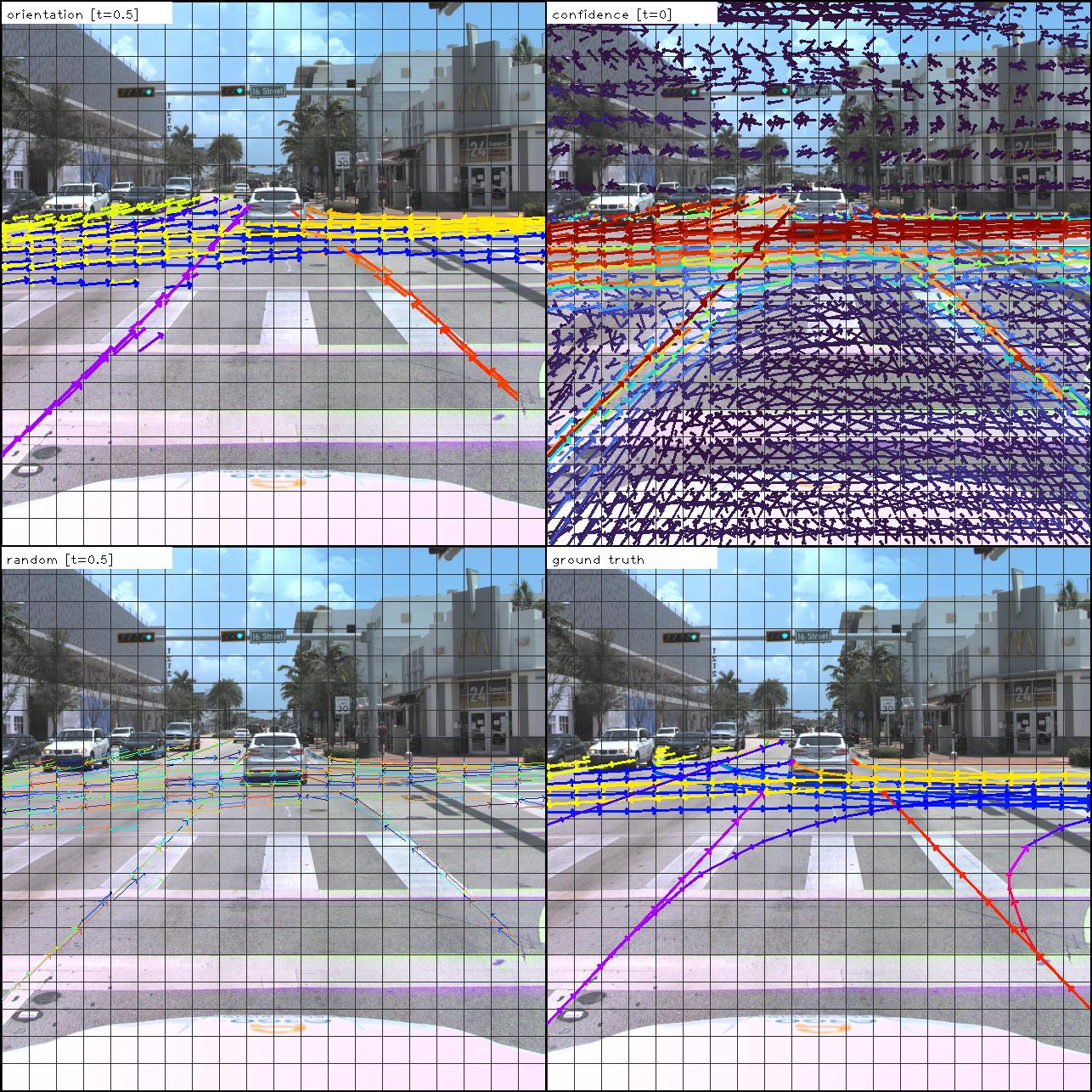}
       
    \end{subfigure}
    \begin{subfigure}[b]{0.49\linewidth}
        \includegraphics[width=\linewidth,trim=464 0 2 576,clip]{figures/test_1bd7db3a-0b42-31cf-ac1a-de88fd9fa721/315969954649927219.jpg__3_pred.jpg}
       
    \end{subfigure}

    \caption{Predictions on the Argoverse dataset. Line segments are color coded according to their orientation in the image.}
    \label{fig:intersection}

\end{figure*}

\autoref{fig:qual_eval_argo} shows centerline predictions on different scenes of the Argoverse dataset with both fixed anchors and the dynamic mapping in training, respectively. 
The selected scenes show the properties of the detection on the Argoverse dataset.
The line segments here are color coded according to orientation. 
This allows to distinguish which lane is in the direction of travel (purple, red, orange) and which is in the opposite direction (yellow). Similarly, \autoref{fig:intersection} shows the prediction of lane borders for an inner city intersection. 

The top row shows a four-lane road with a bicycle lane. 
This is detected by the approach presented here, but is not present in the \ac{GT}.
There are also scenes where the bicycle lane is present in the map and therefore a valid prediction. 

The second scene includes images taken in backlight conditions so that the road is highly reflective. 
Despite this challenge, a mostly correct estimate can still be determined. 

In the third row, shadows and parked vehicles also pose a challenge to the prediction. 
In addition, this is a road with only one direction of travel, so the ego vehicle is in the far left lane. 
This rarely occurs in the training dataset.
The bottom row shows an intersection that was successfully detected. 

The quantitative improvement in results already shown by dynamic mapping without anchors is also evident in the example images.
The polylines are more consistent and have fewer false estimates.

\feedback{Stiller}{Eine zusätzliche nähere Kreuzung wäre cool, zB aus Bild 1. Schließlich ist das in der Einleitung die Motivation, dass auch Komplexe Intersections repräsentierbar sind.
Weiterhin wären Merge Situations, zB 3 nach 2 streifig interessant. Du zeigst zu viele recht einfache Situationen. Weitere komplexe Beispiele machen dein paper interessanter}

\section{Conclusions and Outlook}
\label{cha:conclusions_and_outlook}

In this work, we presented an approach that can detect line-shaped map features in real time. 
Both explicit visible lines (markers) and implicit lines (lane centerlines) can be accurately estimated. 
To this end, the approach employs a purely feed-forward architecture and does not require iterative procedures since the line features are estimated discretized by a cell grid. 
It was shown that the approach is real-time capable, can detect lanes in arbitrary scenes, and is even suitable for complex intersection geometries in urban areas.
\feedback{Stiller}{Das ist mit einer einzigen Intersection wohl kaum für den Leser nachvollziehbar} 
The proposed discretization allows for representing arbitrary directed, intersecting and merging lane geometries in neural networks without explicit modelling. 
To the best knowledge of the authors, this has not been possible in a forward network architecture before.

The presented representation continuously specifies the direction of travel for each line segment and can uniquely distinguish instances of different polylines without resorting to direction discretization or classification.
We have shown the application on lane borders, centerlines, and markings on both highways and urban areas. 
The state of the art of research here predominantly fails already in the representation of intersections. 
Using the classification of each line segment, a wide variety of map features can be identified simultaneously: Markers can be classified according to their semantics (dashed and solid) or centerlines can be distinguished from lane borders and predicted jointly.

In this work, the anchors are determined by $k$-means clustering. 
For future work, it is interesting to investigate whether a more elaborate optimization has advantages over $k$-means. 
This would allow the number of multiple assignments to be minimized directly when the anchors are determined, so that all line segments in the training dataset could be fully mapped.

Related work has already highlighted the lack of suitable line detectors for model-driven intersection estimation. 
So far, these estimators process point-based measurements such as semantic segmentation or image data, as well as line-based features such as trajectories. 
A detector that presents markers or centerlines as polylines properly in real time does not yet exist. 
\feedback{Stiller}{Diesen Satz würde ich nach vorne an den Anfang der Conclusions ziehen:
This contribution closes the gap to detect marked or unmarked lanes by polylines or centerlines in real time.}
The approach presented in this paper has closed this gap.
It provides not only line detections, but simultaneously a hypothesis estimate for the distribution of line segments in the image. 
With model-based estimates, as presented by the author in \cite{meyer2019mcmc}, this representation can be used to probabilistically estimate a complete intersection representation.
An additional \acs{NMS} or connectivity estimation is not even necessary for this purpose. 

\begin{figure*}[tb]
    \centering
    \begin{subfigure}[b]{0.3\linewidth}
        \centering
        \includegraphics[width=1\linewidth,trim=3 769 769 32,clip]{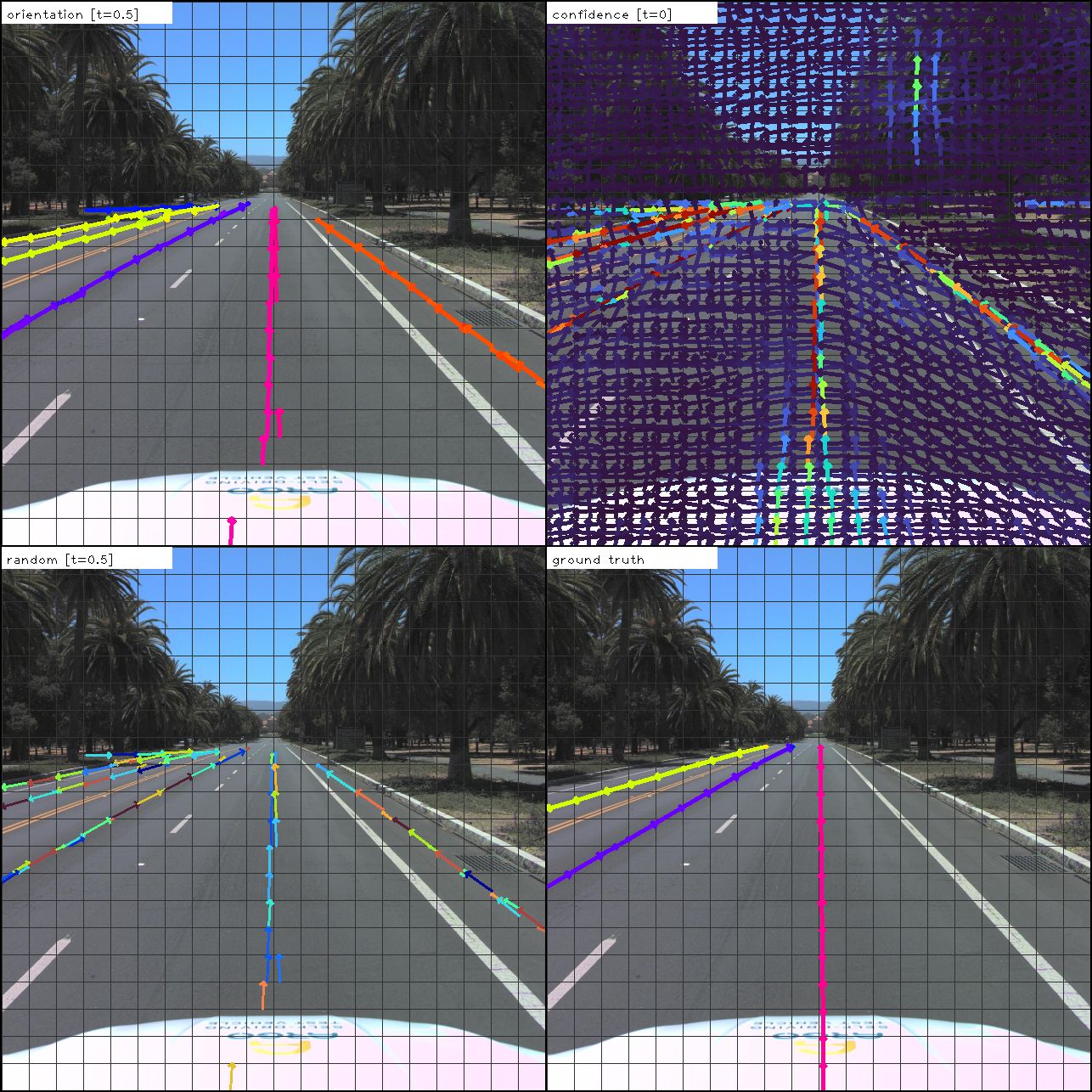}
    \end{subfigure}
    \begin{subfigure}[b]{0.3\linewidth}
        \centering
        \includegraphics[width=1\linewidth,trim=3 769 769 32,clip]{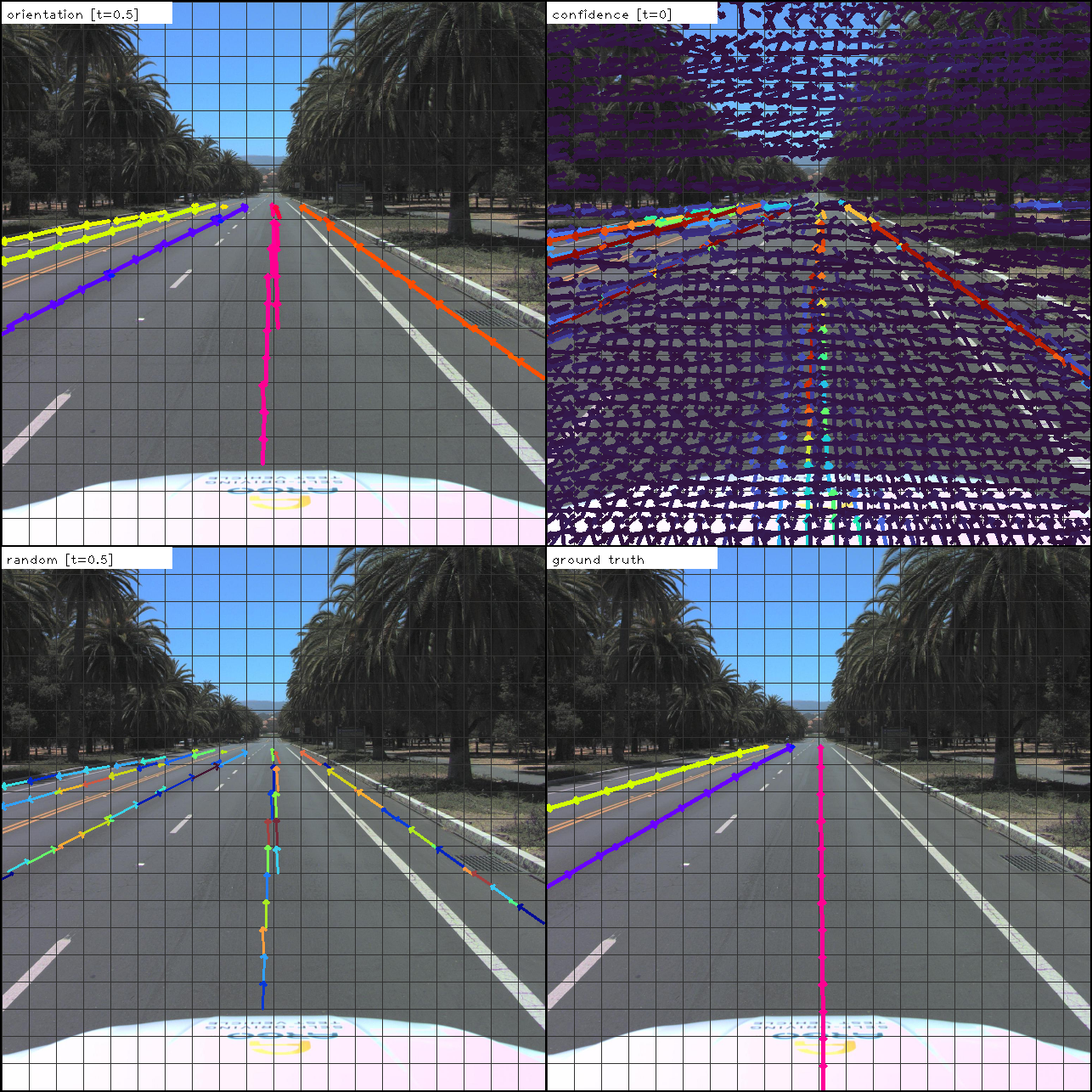}
    \end{subfigure}
    \begin{subfigure}[b]{0.3\linewidth}
        \centering
        \includegraphics[width=1\linewidth,trim=643 128 129 673,clip]{figures/generated/wandb/None_light-sweep-9_epoch-best-val_omega1018_omegageompredconf11_square32_315969765649927217_ep63.png}
    \end{subfigure}
    \begin{subfigure}[b]{0.3\linewidth}
        \centering
        \includegraphics[width=1\linewidth,trim=3 769 769 32,clip]{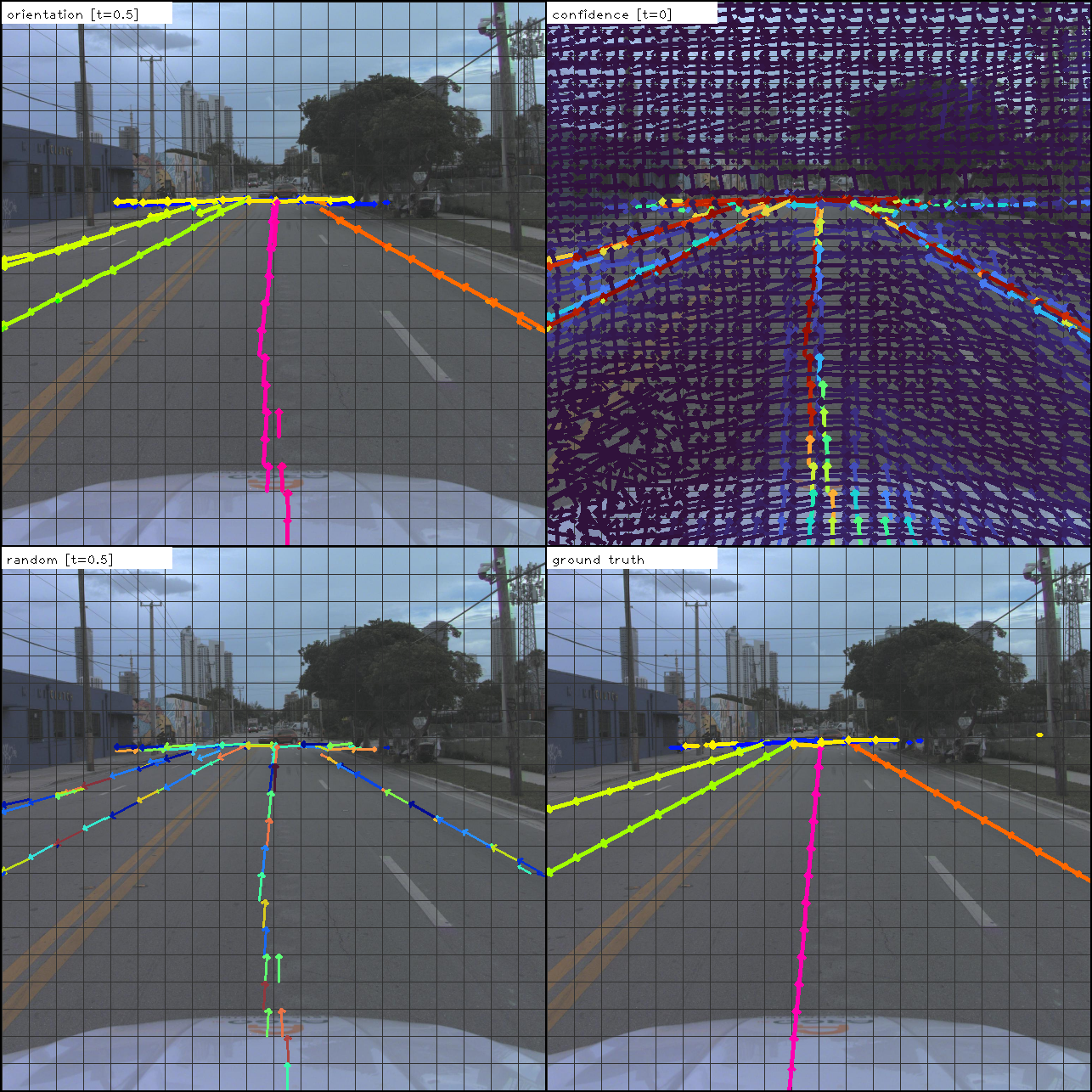}     
        \caption{Anchors}   
        \label{fig:qual_eval_argo_anchor}
    \end{subfigure}
    \begin{subfigure}[b]{0.3\linewidth}
        \centering
        \includegraphics[width=1\linewidth,trim=3 769 769 32,clip]{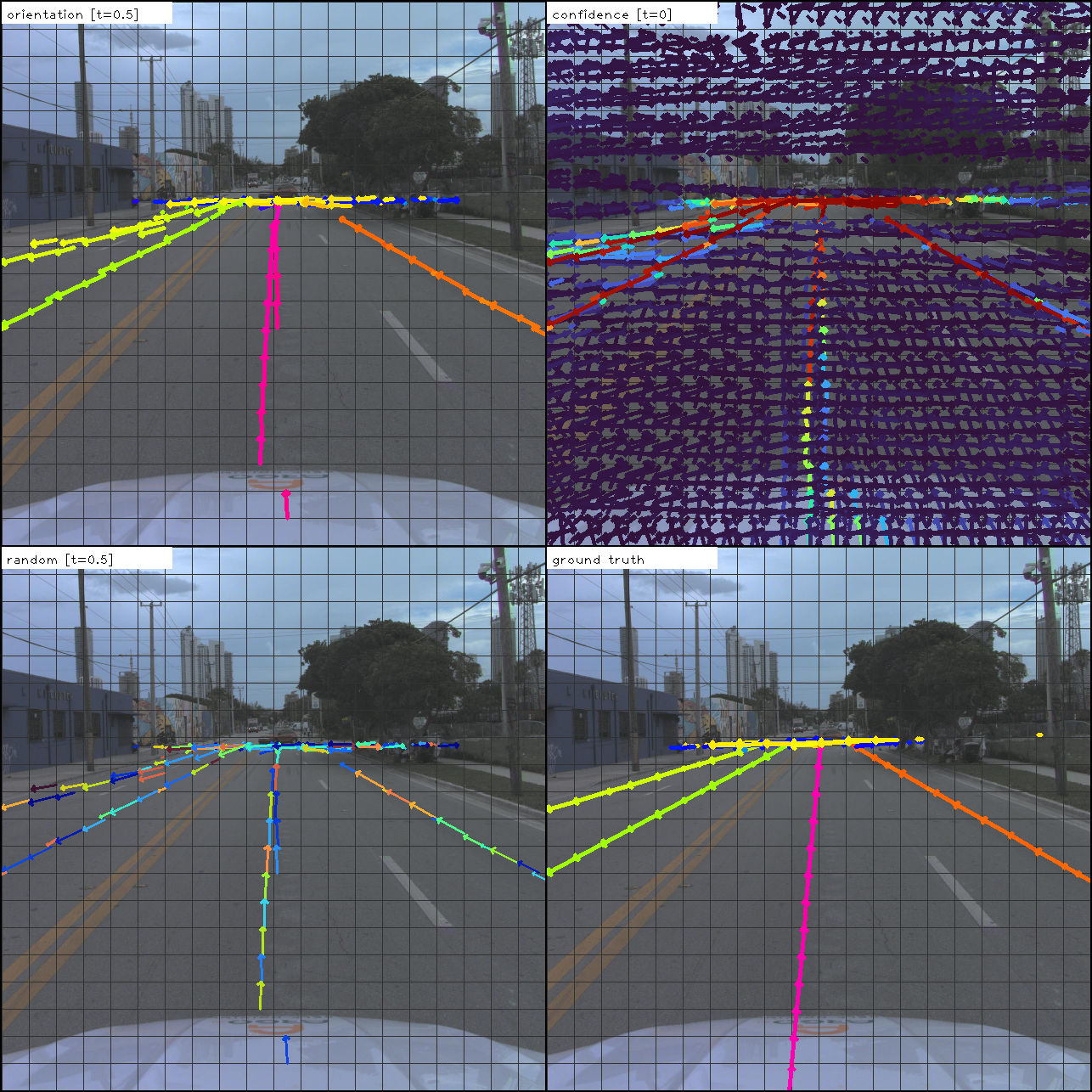}
        \caption{Dynamic}
        \label{fig:qual_eval_argo_dyn}
    \end{subfigure}
    \begin{subfigure}[b]{0.3\linewidth}
        \centering
        \includegraphics[width=1\linewidth,trim=643 128 129 673,clip]{figures/generated/wandb/None_light-sweep-9_epoch-best-val_omega1018_omegageompredconf11_square32_315977901249927221_ep63.png}
        \caption{\acl{GT}}
        \label{fig:qual_eval_argo}
    \end{subfigure}

    \caption{Prediction on Argoverse 2.0~\cite{wilson_ArgoverseNextGeneration_2021} dataset with predefined anchors in column a) and dynamic assignment in column b). The color of the line segments encodes their orientation in the image, giving an impression of the direction of travel.}
    \label{fig:qual_eval_argo}
\end{figure*}

The line segment estimation from this work was designed for application to lanes, but can be used generically for detection of any line entities in images. 
This can enable application not only in mapless driving, but various other disciplines.

Current systems that rely on a high-accuracy map also benefit from this work.  
Semantic localization in a map requires a detector that recognizes elements of the map in the environment so that the detections can be associated with it. 
So far, this has been implemented using area-based estimates such as semantic segmentation.
In \cite{munoz-banon_DALMRRobustLane_2022}, it was shown that line-shaped features can be used to implement fast and highly accurate localization.
This work provides the real-time capable detector for this purpose. 

\bibliographystyle{IEEEtran}
\bibliography{bib/extern_zotero_copy,bib/extern_manually}

\vfill
 
\end{document}